\documentclass[runningheads]{llncs}
\usepackage[T1]{fontenc}
\usepackage{graphicx}
\usepackage{booktabs}
\usepackage[misc]{ifsym}

\usepackage{tabularx} 

\usepackage{mwe}
\usepackage{adjustbox}
\usepackage{hhline} 
\usepackage{cite} 
\usepackage{url}  
\usepackage{subcaption}
\usepackage{booktabs} 
\usepackage[colorlinks=true, linkcolor=black, citecolor=black, urlcolor=black]{hyperref}

\usepackage{algorithm2e}
\usepackage[Table,xcdraw]{xcolor}
\usepackage{xcolor}
\usepackage{physics}
\usepackage{physics, amsmath}
\usepackage{mathtools} 
\usepackage{amssymb} 
\usepackage{multirow}

\def\teal{\textcolor{teal}}

\def\hlinewd#1{%
\noalign{\ifnum0=`}\fi\hrule \@height #1 %
\futurelet\reserved@a\@xhline} 
\begin{document}

\title{DNN-GDITD: Out-of-distribution detection via  Deep Neural Network based Gaussian Descriptor for Imbalanced Tabular Data }

\author{Priyanka Chudasama, Anil Surisetty, Aakarsh Malhotra, Alok Singh\\
\tt\small{\{priyanka.chudasama, anil.surisetty, aakarsh.malhotra, alok.singh2\}@mastercard.com}}
\authorrunning{Chudasama et al.}
%
\institute{AI Garage, Mastercard, India}

\titlerunning{DNN-GDITD}
\maketitle

\begin{abstract}
Classification tasks present challenges due to class imbalances and evolving data distributions. Addressing these issues requires a robust method to handle imbalances while effectively detecting out-of-distribution (OOD) samples not encountered during training. This study introduces a novel OOD detection algorithm designed for tabular datasets, titled \textit{\textbf{D}eep \textbf{N}eural \textbf{N}etwork-based \textbf{G}aussian \textbf{D}escriptor for \textbf{I}mbalanced \textbf{T}abular \textbf{D}ata} (\textbf{DNN-GDITD}). The DNN-GDITD algorithm can be placed on top of any DNN to facilitate better classification of imbalanced data and OOD detection using spherical decision boundaries. Using a combination of Push, Score-based, and focal losses, DNN-GDITD assigns confidence scores to test data points, categorizing them as known classes or as an OOD sample. Extensive experimentation on tabular datasets demonstrates the effectiveness of DNN-GDITD compared to three OOD algorithms. Evaluation encompasses imbalanced and balanced scenarios on diverse tabular datasets, including a synthetic financial dispute dataset and publicly available tabular datasets like Gas Sensor, Drive Diagnosis, and MNIST, showcasing DNN-GDITD's versatility.
\end{abstract}


\section{Introduction} \label{introduction}





In complex domains such as finance \cite{finance}, manufacturing \cite{manufac}, and self-driving vehicles \cite{newsArticle}, numerous safety-critical decisions must be made. Here, spotting unusual patterns in the presence of imbalanced classes is essential, as Out-of-distribution (OOD) points may get confused as minority classes (or vice versa). 
For instance, an autonomous vehicle usually observes more red/green lights vs. a yellow light. However, a self-driving car detected the moon as a yellow light \cite{newsArticle}. Since the model wasn't trained on the moon, moon is an OOD sample. Paying special attention to OOD detection is crucial in these situations. While performing the classification of these datasets in an imbalanced setting for detecting low events such as fraud or defects, the presence of outliers and unforeseen data points can further pose a risk to decision-making models. More specifically, the OOD instances can critically impact the performance of predictive models and lead to sub-optimal outcomes \cite{newsArticle, adv_per2}.


Out-of-distribution (OOD) samples are those data points of some unknown class/distribution that are not seen by the classifier during training \cite{softmax}. On the other hand, in-distribution (ID) samples are data points whose class is seen by the classifier during training. OOD samples involve instances that deviate from ID samples, leading to uncertainties and potential prediction errors, which can be detrimental.

Modern neural/deep networks generalize well when the samples seen during testing are from the same distribution as training \cite{Mdnn4, Mdnn3, Mdnn1, Mdnn2, Mdnn5}. However, they tend to label OOD data as one of the seen classes. Thus, there is a need to accurately classify in-distribution (ID) samples while handling out-of-distribution (OOD) samples during deployment. Existing studies \cite{ softmax,neurips23, lee2018simple, deep_mcdd} perform OOD detection using either (i) an extra OOD class during training, (ii) a confidence-score-based OOD detection (lower confidence prediction for OOD), or (iii) assuming data's distribution and modeling it. These studies seldom operate under imbalanced settings, especially when dealing with tabular data, where research is minimal. In an imbalanced setting, both the minority class and OOD instances scarcity pose a challenge. Existing algorithms \cite{17, 26} struggle to distinguish between OOD and the minority/rare class, leading to potential mis-classifications and reduced model performance. The scarcity of examples from both classes hinders the model's ability to learn distinct features, increasing the risk of confusion between these two inherently different data.

Such classification algorithms should have two properties for imbalanced data: (i) handle classification under an imbalanced setting, and (ii) warn the user or discard predicting suspicious testing samples that are OOD. 
Thus, we propose a classification module for such scenarios, namely DNN-GDITD, that stands for \textbf{D}eep \textbf{N}eural \textbf{N}etwork-based \textbf{G}aussian \textbf{D}escriptor for \textbf{I}mbalanced \textbf{T}abular 
\textbf{D}ata for out-of-distribution data detection. The contributions of the proposed work are as follows:
\begin{enumerate}
    \item Novel loss function to increase inter-cluster distance while creating compact clusters for imbalanced setting
    \item Can be placed over any DNN, as the proposed four loss components can be clubbed with any underlying DNN
    \item Exhaustive experimentation on multiple tabular datasets and algorithms, including softmax-based classification \cite{softmax}, Mahalanobis \cite{lee2018simple}, and Deep-MCDD \cite{deep_mcdd}.
\end{enumerate}
The rest of this paper is organized as follows. We showcase the related work in Section \ref{Related work} and the problem definition in Section \ref{problem definition}. Next, the proposed method is introduced in Section \ref{method}. Section \ref{experiments} presents the experimental setup, results and analysis followed by the ablation of loss term used in DNN-GDITD. Finally, we conclude with the significance of our proposed algorithm in Section \ref{conclude}.

\section{Related Work} \label{Related work}

Out of Distribution (OOD) detection methods can be broadly classified into three categories:
\begin{enumerate}
    \item Considering an extra OOD class during training
    \item Confidence score based OOD detection
    \item Assuming data's distribution and then modeling it.
\end{enumerate}

\noindent\textbf{Extra OOD class during training:} In these OOD methods such as  ~\cite{neurips23},~\cite{ODIN} an extra class is added as an OOD class during training. The extra class contains sample points different from instances of $k$ considered classes. Here, the classification problem becomes a normal classification problem with $k+1$ classes during training and testing. Thus, any classification algorithm can be used if an additional OOD class is added during training. However, the sample count of OOD should be comparable to the count of each $k$ class of interest; otherwise, the OOD pattern may not get captured. Liang and Srikant \cite{ODIN} proposed the ODIN algorithm for OOD detection. ODIN does not require changing a pre-trained neural network to classify pre-defined $k$ classes. It uses temperature scaling and adds small perturbations to the input to separate the softmax score distributions between ID and OOD instances. ODIN used the OOD class to tune hyper-parameters to decide the perturbation intensity and temperature scaling using a validation set containing the OOD class. 

\noindent\textbf{Confidence-score based OOD detection:} Hendrycks and Gimpel \cite{softmax} created a baseline for finding OOD samples during training. Such methods do not require an OOD class during training but instead rely on a confidence score. They used softmax probability scores to classify in and out-of-distribution samples. Their method is based on the assumption that correctly classified examples have higher softmax probabilities than incorrectly classified and OOD examples. Unfortunately, DNN tends to have high confidence even on samples they have never seen before \cite{17, 26}. Further, Guo et al. \cite{guo2017calibration} experimentally show the need for calibrating DNNs as they tend to have high confidence and less accuracy. 

As an alternative, K. Lee et al. \cite{lee2018simple} use Mahalanobis distance-based confidence score with respect to the closest class-conditional distribution out of $k$ classes. Their algorithm applies to any pre-trained softmax neural classifier (without re-training) for detecting OOD samples. However, as K. Lee et al. \cite{deep_mcdd} explains, DNNs trained using softmax are not optimal in distinguishing OOD samples from in-distribution (ID) samples. Works such as \cite{softmax4, softmax1,softmax3,softmax2} show that prediction probability from a softmax distribution has a poor direct correspondence to confidence. Softmax-based classifiers tend to overlap significantly between difficult ID and OOD samples. Difficult ID samples are those whose probability value tends to be same for all of the classes, making the model less confident in it's prediction, same is seen for OOD samples. For this reason, there is less gap between confidence scores of an ID vs an OOD sample. 

\noindent\textbf{Assuming data's distribution and modeling it:} 
In this approach, just like the previous approach we do not require OOD samples during training. Here, the assumption is rather made on the distribution of training classes. For instance, D. Lee et al. \cite{deep_mcdd} tried to fit training data into Gaussian spheres and proposed Deep-MCDD to detect OOD samples and label ID samples using spherical decision boundaries. Softmax-based classifiers have linear decision boundaries to separate ID and OOD samples. Deep-MCDD transforms the per-class embedding distribution to spherical decision boundaries that consider all class conditional samples as independent Gaussian distributions. Any data point is marked as ID or OOD based on its distance from any cluster. They add the KL Divergence between the class conditional probability and a Gaussian distribution to convert the former into the latter, given as:
\begin{equation}\label{kl}
    KL(\mathbf{P}_i||\mathcal{N}(\mu_i, \sigma_i) ) \mbox{ where } \mathbf{P}_i = \frac{1}{N_i} \sum_{y_j = i} \delta(e - f(x_j))
\end{equation}
Here, the Gaussian distribution is denoted by $\mathcal{N}(\mu_i, \sigma_i)$, where $\mu_i$ and $\sigma_i$ represent mean and standard deviation for class $i$, respectively. The equation above shows that Deep-MCDD assumes the class conditional distribution in the embedding space to be discrete. 
For any vector $e$ in embedding space, its probability value is non-zero only when it is an embedding vector of a training sample. Moreover, the class conditional probability value equals the count of training samples with embedding vector as $e$ by a total number of training samples in the conditional class. The problem with such an assumption is that 
\begin{enumerate}
    \item It does not consider the distance between two embedding vectors and assigns the same probability if the frequency of two embedding vectors is the same.
    \item For minority classes, the number of data points is less. Consequently, each embedding vector of training samples in the minority class gets assigned a high probability. Furthermore, for a given class it's distribution in the embedding space is continuous , and for continuous distribution, the probability of seeing one value from the possible values $x$ can take is always zero. Thus, contradicting the choice of probability distribution assumed for each class in equation \ref{kl}.
\end{enumerate}

Further, one does not always have access to OOD samples while training. Moreover, it can never be exhaustive, even if one has access to OOD samples. Consequently, we provide a solution for detecting OOD samples in imbalanced settings under the category 2 and 3. Moreover, to solve the issues as pointed out in various methods for OOD detection above, we propose Out-of-distribution detection via \textbf{D}eep \textbf{N}eural \textbf{N}etwork based \textbf{G}aussian \textbf{D}escriptor for \textbf{I}mbalanced \textbf{T}abular \textbf{D}ata (DNN-GDITD). Our algorithm is based on the OOD detection method, which assumes that OOD samples are not seen during training. Hence, we do not compare our algorithm with OOD methods based on considering an extra OOD class during training. Furthermore, the novelty in our methods comes from the decision boundaries created to separate each class cluster and not from the inherent model being used. Thus, we compare our proposed method DNN-GDITD with similar classification modules like softmax \cite{softmax}, Mahalanobis-based confidence score \cite{lee2018simple}, and Deep-MCDD \cite{deep_mcdd}. Consequently, our method can be set up on top of any base DNN classifier to improve OOD detection.

\section{Problem Definition} \label{problem definition}
Let  $\{X, Y\}$ be the collection of all samples seen during training. We assume that there are $k$ distinct classes. Thus, for $x \in X$ we have it's corresponding label, $y_x \in \{1,2, \ldots, k\}$. Further, let the entire training and testing dataset be denoted as $\mathcal{D} \coloneqq \mathcal{D}_{train} \cup \mathcal{D}_{test}$. Here, $\mathcal{D}_{train} =  \{X, Y\}$ with $k$ distinct classes and  $\mathcal{D}_{test} = \{X', Y'\}$, where the predictions of input samples $X'$ are evaluated against labels $Y'$. However, the $\mathcal{D}_{test}$ may contain samples from either the $k$ classes (seen during training) or from a distribution not seen during training, called an OOD class. Further, the considered $k$ classes are imbalanced. Having an imbalanced class during the training makes the problem of OOD detection during testing more challenging. In such a setting, the classifier is supposed to rightly assign the sample $x$ into one of the $k$ imbalanced classes or label OOD when $y_x \not\mathrel{\epsilon} \{1,2, \dots, k\}$. Thus, the aim is to propose a classification module that can work in an imbalanced setting and efficiently classify any testing sample into either one of the $k$ classes or as an OOD sample. 


\section{Proposed Algorithm: DNN-GDITD} \label{method}
 Consider a training sample $x$ $\in$ $X$ with corresponding ground-truth $y_x \in Y$. OOD detection via DNN-GDITD is a classification module that takes embedding vectors $f(x)$ from a DNN $f(:,W) \in \mathbb{R}^d$, with learnable parameters $W$. Inspired by Gaussian Discriminant Analysis, the proposed DNN-GDITD tries to transform the embedding space into a collection of independent Gaussian-distributed clusters (spheres) for $k$ classes seen during training with spherical decision boundaries. Consequently, each cluster $ i \in \{1,2, \ldots, k\}$ is assumed to have a mean $\mu_i$ and standard deviation $\sigma_i$.

Suppose we're given a training instance $x$ sampled from mini-batch $B$ from $\mathcal{D}_{train}$. We define its distance from the $i^{th}$ cluster for $ i \in \{1,2,\ldots k\}$ in the latent space using the $i^{th}$ class conditional probability as:
\begin{equation}\label{eq2}
    D_i(x) \coloneqq -log \; Pr(x|i)
\end{equation}
 Further, we assume that data in latent space is distributed as $k$ independent Gaussian spheres. Thus $Pr(x|i) \approx \mathcal{N}(\mu_i,\sigma_i)$ for $ i \in \{1,2,\ldots k\}$. Hence,
 \begin{equation}
     D_i(x) = -log\; \mathcal{N}(f(x)|\mu,\Sigma),
 \end{equation}
 where $\mu = [\mu_1, \mu_2, \ldots \mu_k]$ and $\Sigma = [\sigma_1, \sigma_2, \ldots \sigma_k]*I$. Here, $I_{k\times k}$ is the identity matrix. Consequently, the distance from a given training data point $x$ from the $i^{th}$ cluster in the latent space can be defined as: 
 \begin{equation}\label{prob}
     D_i(x) =  \frac{\norm{f(x) - \mu_i}^2}{2\sigma_i^2} + log(\sigma_i)^d, 
 \end{equation}
where $d$ is the dimension of the latent space.
 
Using the above-mentioned class-specific distance, the objective for the proposed DNN-GDITD algorithm is to:
\begin{enumerate}
    \item Transform  the latent space into a $k$ independent Gaussian-distributed clusters (spheres). 
    \item Create compact and well-separated clusters.
    \item Define a measure that can handle imbalanced data and can identify OOD samples during testing.
\end{enumerate}
Consequently, we propose four losses, as described in the upcoming subsections.

\begin{figure}[ht]
    \centering
    \begin{subfigure}{0.48\textwidth} 
        \includegraphics[trim={28cm  15.5cm 27cm 10.5cm},clip,width=\linewidth,height=5cm]{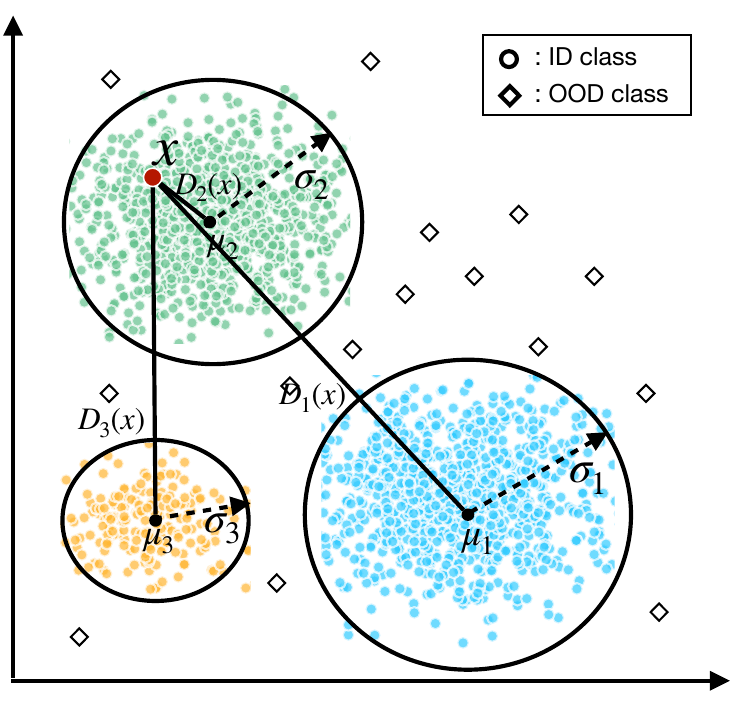} 
        \caption{$x \in$ ID class : $\zeta_2 >0$ and $\zeta_1, \zeta_3 <0$} 
        \label{fig:plot_1} 
    \end{subfigure}
    \hfill 
    \begin{subfigure}{0.48\textwidth} 
        \includegraphics[trim={30cm  15.5cm 25cm 10.5cm},clip,width=\linewidth,height=5cm]{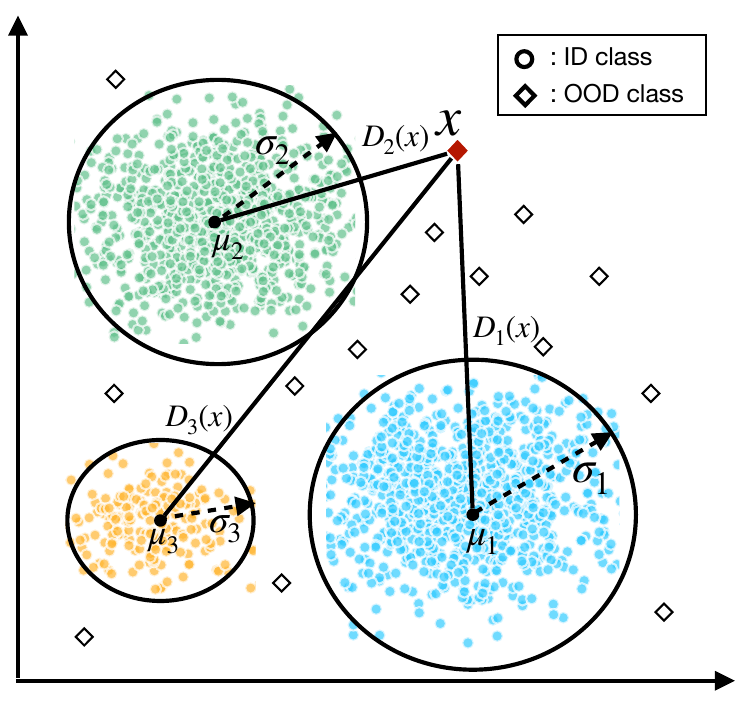} 
        \caption{$x \in$ OOD class : $\zeta_1, \zeta_2, \zeta_3 <0$} 
        \label{fig:plot_2} 
    \end{subfigure}
    \caption{Sign of score $\zeta_i(x) = \sigma_i - D_i(x)$ with respect to different clusters $i \in \{1,2,3\}$ when $x$ is an ID sample (case (a)) vs when x is an OOD sample (case (b)). $(\mu_i, \sigma_i)$ represent each clusters centre and radius respectively. Score wrt the true class should be positive, whereas score wrt the rest of classes should be negative. Thus, in case (b) as $x$ is an OOD sample as $\zeta_i(x) <0$ for all clusters.  }
    \label{fig:plots} 
\end{figure}

For a training instance $x$ with label $y_x$, we want its distance from its own cluster $(\mu_{y_x}, \sigma_{y_x})$ to be less and more from other clusters (refer to Fig. \ref{fig:plot_1} (a)). $(\mu_i, \sigma_i)$ can be considered analogous to the cluster's center and radius, respectively. To this extent we define score $(\zeta)$ for each cluster as follows, 
\begin{equation}\label{score}
    \zeta_i(x) \coloneqq \sigma_i - D_i(x) \hbox{ for } i \in {1,2, \ldots k},
\end{equation}

 Eventually we want, $\zeta_{y_x}(x) \geq 0 $ , that is data point $x$ should lie inside the cluster $y_x$. Data point $x$ should lie outside rest of the clusters, i.e., $\zeta_{y_x}(x) < 0 $ for $i \neq y_x$. Thus, we will have loss functions which decrease intra-class distance while making sure that $\zeta_i(x)$ satisfies the aforementioned conditions.
\RestyleAlgo{ruled}

\SetKwComment{Comment}{$\triangleright$ }{}

\begin{algorithm}[hbt!] \footnotesize 
\caption{Pseudo code for DNN-GDID}\label{alg:two}
\KwData{Training (in-distribution) data: $\mathcal{D}_{train} = \{X,Y_X\}$ with $k$ classes, testing data $\mathcal{D}_{test} = \{X',Y_{X'}\}$ with OOD and k classes ( as seen during training), base DNN: $f(:,W) \in \mathbb{R}^d$, mean and standard deviation: $(\mu_i, \sigma_i)$ for $i \in \{1, 2, \ldots, k\}$ for k class clusters in latent space.}
\For{iter  $ \in \{1,2, \dots\}$ }
   {sample a mini-batch $B$ from $\mathcal{D}_{train}$\;
   \For{$x \in B:$}
   { \Comment{\teal{feed sample point into DNN to get vector representation in latent space:}}
   $f(:,W): x \mapsto f(x)$\; 
   \Comment{\teal{Get distance of x from each k class cluster after converting the k-classes distribution in latent space obtained from $f(:,W)$ to isotropic Gaussian distributions:}}
   $D_i(x)= \frac{\norm{f(x) - \mu_i}^2}{2\sigma_i^2} + log(\sigma_i)^d, 1\leq i \leq k$ 
    \Comment{\teal{Predicted class for x:}}
    $\tilde{y}_x = \underset{1\leq i \leq k}{\mathrm{argmax}} (\zeta_i(x))$\;
    where, $\zeta_i(x) \coloneqq \sigma_i + D_i(x) $ 
   }
   \Comment{\teal{Pull loss: to reduce the distance from sample's own class and make compact clusters, }}
   $L_{\mathrm{P}} = \sum_{x \in B} D_{y_x}(x)$\; 
   \Comment{\teal{Score loss: to make sure score from sample's own class is non-negative and from rest classes it's negative, }}

 $L_{\mathrm{SL}} =
    \begin{cases}
    \sum_{x \in B} \sum_{i} \frac{\exp(\zeta_i(x))}{\#|B|} \hbox{ for } i \in \{1,2,\ldots, k\}\backslash \{y_x\}, \\
        \sum_{x \in B}\left( \mathrm{ReLU} (- \zeta_{y_x}(x)) + \mathrm{log}(1 +\zeta_{y_x}(x)^2 ) \right), \hbox{else}
    \end{cases}$\;
    
   \Comment{\teal{Distance and Score-based Effective Focal loss: }}
   $L_{\mathrm{EFL_1}} = \sum_{x \in B}\mathrm{EFloss}(y_x^o, 1/D(x))$\; \Comment{\teal{$y_x^o$ is one hot label vector for $x$}}
   $L_{\mathrm{EFL_2}} = \sum_{x \in B}\mathrm{EFloss}(y_x^o,\zeta_i(x))$\;
   \Comment{\teal{Net-loss}}
   $\mathcal{L}_{net}= L_{P} + L_{SL} + L_{EFL_1} + L_{EFL_2}$\;
   \Comment{\teal{Use Block Coordinate Descent (BCD) to update:}}
    (a) DNN's parameters viz $W$ and\;
   (b) Mean and standard deviation for k in-distribution classes viz $(\mu_i, \sigma_i)$ for $i \in \{1, 2, \ldots, k\}$\;
   }
   \Comment{\teal{Prediction for test data containing both in-distribution (ID) and OOD samples:}}
   \For{$x' \in \mathcal{D}_{test}$}
   {\begin{equation*} 
  \tilde{y}_x' =
    \begin{cases}
       OOD & \textbf{if } \zeta_i(x) < 0 \hbox{ for } i \in \{1,2, \ldots, k\}\\
      \underset{1\leq i \leq k}{\mathrm{argmax}} (\zeta_i(x)), & \textbf{otherwise}
    \end{cases}       
\end{equation*}}
\end{algorithm}
\label{our_algorithm}  

\subsection{Pull loss}
Let us consider a training instance $x$ with ground truth label $y_x$, sampled from mini-batch $B \in \mathcal{D}_{train}$. Then, we define pull loss as follows:
\begin{equation}\label{pull_loss}
    L_{\mathrm{p}} = \sum _{x \in B}D_{y_x}(x) = \sum _{x \in B}\frac{\norm{f(x) - \mu_{y_x}}^2}{2\sigma_{y_x}^2} + log(\sigma_{y_x})^d
\end{equation}
Pull loss aims to reduce the distance of a given training instance $x$ from its original class $y_x$ and reduce the $\sigma_{y_x}$, thus making compact clusters.

\subsection{Scores based loss (SL)}
We want $\zeta_{y_x}(x) \geq 0 $ and $\zeta_{i}(x) < 0 $ for $i \neq y_x$. To achieve it, we create a score-based loss. When $i \neq y_x$, we want to have a negative score from the rest of the classes ($\zeta_{i}(x)<0$). The loss $L_{\mathrm{SL}}$ in this case takes the exponential of score $\zeta_{y_x}(x)$, ensuring that scores are negative for the rest of the classes. Similarly, when $i = y_x$, we want the score to be positive. Thus, we feed in the negative scores through ReLU activation and also take the log of scores to ensure we get smaller values for scores. Essentially, the score loss is defined as: 
{\begin{equation}\label{sl_loss}
 L_{\mathrm{SL}} =
    \begin{cases}
    \sum_{x \in B} \sum_{i} \frac{\exp(\zeta_i(x))}{\#|B|} \hbox{ for } i \in \{1,2,\ldots, k\}\backslash \{y_x\}, \\
        \sum_{x \in B}\left( \mathrm{ReLU} (- \zeta_{y_x}(x)) + \mathrm{log}(1 +\zeta_{y_x}(x)^2 ) \right), \hbox{ else}
    \end{cases}       
\end{equation}}


\subsection{Effective focal loss (EFL)}
Lastly, to tackle classification in an imbalanced setting, we use weighted focal loss. Focal loss \cite{focal} is usually used for object detection and in classification \cite{effective}. It helps to emphasize difficult examples during backpropagation. Further, Cui et al. \cite{effective} introduced class-balanced loss using an effective number for class imbalance classification. We use this effective number to create a weighted focal loss, which we call \textit{effective focal loss }(EFL). Cui et al. \cite{effective} measure data overlap by associating a small neighboring region with each sample rather than a single point. Thus, \textit{effective focal loss} for sample $x$ with actual class $y_x$ and corresponding class's predicted probability $p_{y_x}(x)$ is defined as:
\begin{equation} \label{focal_loss}
   \mathrm{EFloss}(y_x, p_{y_x}) \coloneqq  \frac{1-\beta}{1-\beta^{n_{y_x}}} (1-p_{y_x}(x))^{\gamma} log(p_{y_x}(x))
\end{equation}
Here, $\gamma$ is the focus parameter. Increasing $\gamma$ increases the focus on difficult sample learning. $\gamma = 0$ will give us the usual cross-entropy loss.  The effective number, $\beta$ is usually considered as $1/\#|B|$, where $n_{y_x}$ is the number of samples of class $y_x$ in current batch $B$. We use EFL as defined on both $D(x)$ and $\zeta_i(x)$ as follows:
\begin{align} 
  L_{\mathrm{EFL_1}} &= \sum_{x \in B}\mathrm{EFloss}(y_x^o, 1/D(x)). \label{efl1_loss} \\
  L_{\mathrm{EFL_2}} &= \sum_{x \in B}\mathrm{EFloss}(y_x^o,\zeta_i(x)). \label{efl2_loss}
\end{align}

Collectively, the net loss for the proposed DNN-GDITD algorithm is given as:
\begin{equation}\label{net_loss}
\mathcal{L}_{net}=L_{p}+L_{SL}+L_{\mathrm{EFL_1}}+L_{\mathrm{EFL_2}}
\end{equation}

Post training, the label $\tilde{y}_x$ for a testing data point $x$ is predicted as per the maximum argument of $\zeta_i(x) \coloneqq \sigma_i - D_i(x)$.  If  $\zeta_i(x)$ is negative for all $i \in \{1,2, \ldots, k\}$, $x$ will be marked as OOD, else $x$ will be ID and the maximum argument of $\zeta_i(x)$ will be chosen as $\tilde{y}_x$. (Refer to Fig. \ref{fig_plots} for better understanding and visualization.) Equivalently, we have
   {\begin{equation}\label{prediction}
  \tilde{y}_x' =
    \begin{cases}
       OOD & \textbf{if } \zeta_i(x) < 0 \hbox{ for } i \in \{1,2, \ldots, k\}\\
      \underset{1\leq i \leq k}{\mathrm{argmax}} (\zeta_i(x)), & \textbf{otherwise}
    \end{cases}       
\end{equation}}

For ease of understanding of DNN-GDITD, we provide pseudo-code for our algorithm in \ref{our_algorithm}.


\section{Experiments details and Analysis} \label{experiments}
\subsection{Datasets}\label{datasets}

We show the efficacy of DNN-GDITD on four different tabular datasets: (i) Synthetic financial, (ii) Gas Sensor \cite{data_gas}, (iii) Drive Diagnosis \cite{data_drive}, and (iv) MNIST \cite{data_mnist}. Table \ref{stat} gives all the datasets' details. These four datasets are selected to showcase the efficacy of the proposed work in all three scenarios: highly imbalanced, imbalanced, and balanced. We use the synthetic financial dataset to showcase results in highly imbalanced settings. The synthetic financial dataset is created from a fraud dataset using a modified SMOTE \cite{privacy} technique to preserve privacy. It consists of three labels: legitimate-dispute, first-party fraud (FPF), and third-party fraud (TPF), with FPF:TPF:legitimate-dispute ratio being $08:46:46$. Thus, the synthetic financial dataset falls under a highly imbalanced data category. 
For all other publicly available datasets mentioned above, we consider the OOD class as class 0. The minority class is considered without loss of generality as class 1 for the Drive and MNIST datasets as they are balanced. Gas Sensor is an imbalanced data with a minority class as class 2. \\
Further, to show the efficacy of OOD detection in an imbalanced setting, we introduce different levels of imbalance into the data using \textit{Minority-class Down-Sampling Ratios} (MDSR). MDSR indicates the percentage of data to be considered from the considered minority class at random to create $\mathcal{D}$ for training and testing. 
We use different MDSR values ranging from 1, representing zero down-sampling, to 0.10, meaning that only 10\% of minority class data originally present will be used during training as a part of ID class. Further, as Synthetic financial data is highly imbalanced we use MDSR values of $1, 0.3$ and $0.25$ only, so that enough samples of minority class are seen per batch during training.

\begin{table}[hbt!]
\centering
\resizebox{12.25cm}{!}{
\begin{tabular}{|c|cc|cc|}
\hline
\multirow{2}{*}{\textbf{Dataset}} & \multicolumn{2}{c|}{\textbf{Imbalanced}} &  \multicolumn{2}{c|}{\textbf{Balanced}}  \\ \cline{2-5} 
& \multicolumn{1}{c|}{\begin{tabular}[c]{@{}c@{}}Synthetic Financial \end{tabular}} & \multicolumn{1}{c|}{\begin{tabular}[c]{@{}c@{}}Gas Sensor \cite{data_gas}\end{tabular}} &  \multicolumn{1}{c|}{ \begin{tabular}[c]{@{}c@{}}Drive  Diagnosis \cite{data_drive}\end{tabular}} & \multicolumn{1}{c|}{ \begin{tabular}[c]{@{}c@{}}MNIST  \cite{data_mnist}\end{tabular}} \\ \hline\hline

\# Features & \multicolumn{1}{c|}{943} & \multicolumn{1}{c|}{128}  & \multicolumn{1}{c|}{48} & \multicolumn{1}{c|}{784}\\ \hline

\# Samples  & \multicolumn{1}{c|}{143,670} & \multicolumn{1}{c|}{30,000} & \multicolumn{1}{c|}{ 120,000} & \multicolumn{1}{c|}{70,000} \\ \hline

\# Classes  & \multicolumn{1}{c|}{3} & \multicolumn{1}{c|}{6} & \multicolumn{1}{c|}{ 11} & \multicolumn{1}{c|}{10}\\ \hline

ID classes & \multicolumn{1}{c|}{FPF , TPF} & \multicolumn{1}{c|}{classes 1 to 5} & \multicolumn{1}{c|}{ Classes 1 to 10} & \multicolumn{1}{c|}{Classes 1 to 9} \\ \hline

OOD class  & \multicolumn{1}{c|}{Legitimate-dispute} & \multicolumn{1}{c|}{class 0}  &  \multicolumn{1}{c|}{class 0} & \multicolumn{1}{c|}{class 0} \\ \hline

\begin{tabular}[c]{@{}c@{}}Minority class\end{tabular} & \multicolumn{1}{c|}{FPF} & \multicolumn{1}{c|}{class 2} & \multicolumn{1}{c|}{ None}  & \multicolumn{1}{c|}{ None}\\ \hline
\end{tabular}
}
\vspace{0.2cm}
\caption{Statistics of imbalanced and balanced tabular datasets used for experimental comparison of algorithms.}
\label{stat}
\end{table}

\subsection{Implementation Details}

We follow the original settings proposed by the authors in respective baselines \cite{softmax, lee2018simple, deep_mcdd}. For our experiments, we used $128$ as the latent dimension size in all 3-layers of the MLP network. We used the Adam optimizer with a learning rate of $0.001$ and a batch size of $200$ for all four datasets for Softmax \cite{softmax} and Mahalanobis \cite{lee2018simple}. We use Block Coordinate Descent (BCD) \cite{bcd} with a batch size of $200$ for Deep-MCDD \cite{deep_mcdd} and DNN-GDITD (our) algorithm. This is because updating DNN's parameters $W$ and $(\mu_i, \sigma_i)$ together (for $i \in \{1, 2, \ldots, k\}$) is difficult via popular usual gradient descent methods like SGD or Adam. Thus, we use BCD for optimizing and updating the parameters of our model. More specifically, BCD alternatively updates $W$ and $(\mu_i, \sigma_i)$ while fixing the other set of parameters to minimize the loss function. Before feeding our data into our module, we apply z-score normalization on all the data so that data follows the standard normal distribution. We ran each algorithm for a maximum of $100$ epochs with 5-fold cross-validation. We give the same weightage to each loss while calculating net loss as defined in equation \ref{net_loss} with the $\gamma$ parameter used in focal loss \ref{focal_loss} is set to 1. The best results from each algorithm is mentioned in Table \ref{table_class}. The results can also be visualized with varying MDSR in Fig. \ref{fig_plots}.

\subsection{Results and Analysis} \label{evaluation}
The test data comprises ID and OOD samples. In Table \ref{table_class}, we report classification accuracy for the ID classes and AUPR for minority classes at various MDSR rates. Since the minority class's data is less compared to other ID classes, reporting classification accuracy for the minority class will not help in justifying a model's performance. Thus, we report AUPR values from various models for the minority class. Furthermore, for OOD samples seen during testing, we show the efficacy of our algorithm DNN-GDITD using three metrics: TNR at $85\%$ TPR, AUROC, and AUPR score as the OOD class has almost as many samples as the rest of the ID classes. This helps us validate our model on all the aspects of OOD detection.

\begin{figure}[ht] 
\includegraphics[width = \textwidth]{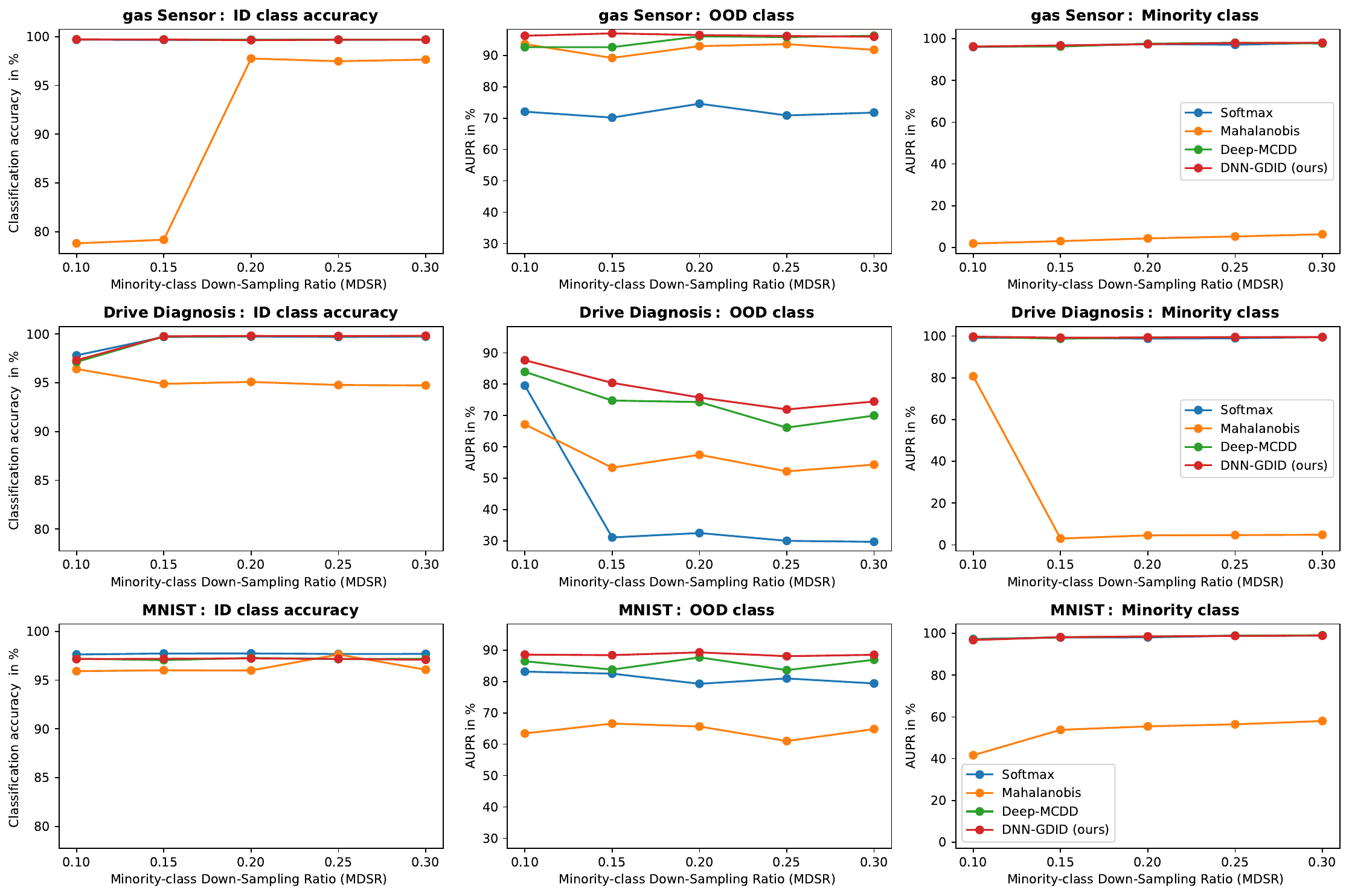}
\centering
\caption{Graphical comparison of Softmax \cite{softmax}, Mahalanobis \cite{lee2018simple}, Deep-MCDD \cite{deep_mcdd} vs DNN-GDITD (ours) on publicly available dataset Gas Sensor \cite{data_gas}, Drive Diagnosis \cite{data_drive} and MNIST \cite{data_mnist}.} \label{fig_plots}
\end{figure}

Further, we have divided Table \ref{table_class} into two parts, one for imbalanced datasets in Table \ref{table_class}(a) and the other for balanced datasets in Table \ref{table_class}(b). This helps us understand the efficacy of our algorithm on both balanced and unbalanced settings separately. Consequently, during training, the same MDSR value in balanced and unbalanced settings can have different ratios of considered minority class samples compared to the rest of ID class samples. To visually showcase the proposed DNN-GDITD algorithm's efficacy, we show classification accuracy plots for ID data, AUPR for OOD, and AUPR for minority classes at various MDSR rates in Fig. \ref{fig_plots}. We can observe from Fig. \ref{fig_plots} that the proposed DNN-GDITD performs comparably in classifying ID classes, with the best value obtained using Softmax. This is seen as our algorithm DNN-GDITD is conservative while marking a sample as ID class. This behaviour can be justified by the score \ref{score} assignment done for each sample. Since we consider a radius of the cluster as the standard deviation of the cluster, we end up marking only the highly confident samples as one of ID classes and rest are marked as OOD class. This approach helps us have a low false positive while marking any test sample as one of the ID class. A low false positive is always appreciated in many applications including fraud detection, Autonomous self driving, healthcare and  many more. Furthermore, after comparing all three aspects including ID class, OOD class, and minority class performance, it can be observed that, the all-rounded performance is given by the DNN-GDITD (ours) algorithm (represented by red color in Fig. \ref{fig_plots}).
\begin{table}[htbp]
    \centering
    \begin{subtable}{\textwidth}
        \centering
        \resizebox{12.0 cm}{!}
        {
\begin{tabular}{|clc|ccc|cccccc|}
\hline
\multicolumn{3}{|c|}{{\color[HTML]{000000} Dataset}}                                                                                                                                                            & \multicolumn{3}{c|}{{\color[HTML]{000000} \begin{tabular}[c]{@{}c@{}}Synthetic financial \\ Dataset\end{tabular}}}                                              & \multicolumn{6}{c|}{{\color[HTML]{000000} \begin{tabular}[c]{@{}c@{}}Gas Sensor\\ Dataset\end{tabular}}}                                                                                                                                                                                                                                               \\ \hline
\multicolumn{3}{|c|}{{\color[HTML]{000000} \begin{tabular}[c]{@{}c@{}}Minority-class Down-\\ Sampling  Ratio (MDSR)\end{tabular}}}                                                                                       & \multicolumn{1}{c|}{{\color[HTML]{000000} 1}}              & \multicolumn{1}{c|}{{\color[HTML]{000000} 0.3}}            & {\color[HTML]{000000} 0.25}           & \multicolumn{1}{c|}{{\color[HTML]{000000} 1}}              & \multicolumn{1}{c|}{{\color[HTML]{000000} 0.3}}            & \multicolumn{1}{c|}{{\color[HTML]{000000} 0.25}}           & \multicolumn{1}{c|}{{\color[HTML]{000000} 0.2}}            & \multicolumn{1}{c|}{{\color[HTML]{000000} 0.15}}           & {\color[HTML]{000000} 0.1}            \\ \hline \hline
\multicolumn{2}{|c|}{{\color[HTML]{000000} }}                                                                                                  & {\color[HTML]{000000}  \cite{softmax}}                                 & \multicolumn{1}{c|}{{\color[HTML]{000000} \textbf{99.45}}} & \multicolumn{1}{c|}{{\color[HTML]{000000} \textbf{99.41}}} & {\color[HTML]{000000} \textbf{99.43}} & \multicolumn{1}{c|}{{\color[HTML]{000000} \textbf{99.67}}} & \multicolumn{1}{c|}{{\color[HTML]{000000} \textbf{99.70}}} & \multicolumn{1}{c|}{{\color[HTML]{000000} 99.68}}          & \multicolumn{1}{c|}{{\color[HTML]{000000} \textbf{99.68}}} & \multicolumn{1}{c|}{{\color[HTML]{000000} 99.67}}          & {\color[HTML]{000000} 99.68}          \\ \cline{3-12} 
\multicolumn{2}{|c|}{{\color[HTML]{000000} }}                                                                                                  & {\color[HTML]{000000}  \cite{lee2018simple}}                             & \multicolumn{1}{c|}{{\color[HTML]{000000} 97.68}}          & \multicolumn{1}{c|}{{\color[HTML]{000000} 98.04}}          & {\color[HTML]{000000} 92.08}          & \multicolumn{1}{c|}{{\color[HTML]{000000} 90.65}}          & \multicolumn{1}{c|}{{\color[HTML]{000000} 97.65}}          & \multicolumn{1}{c|}{{\color[HTML]{000000} 97.48}}          & \multicolumn{1}{c|}{{\color[HTML]{000000} 97.76}}          & \multicolumn{1}{c|}{{\color[HTML]{000000} 79.18}}          & {\color[HTML]{000000} 78.81}          \\ \cline{3-12} 
\multicolumn{2}{|c|}{{\color[HTML]{000000} }}                                                                                                  & {\color[HTML]{000000}  \cite{deep_mcdd}}                               & \multicolumn{1}{c|}{{\color[HTML]{000000} 99.31}}          & \multicolumn{1}{c|}{{\color[HTML]{000000} 99.21}}          & {\color[HTML]{000000} 99.22}          & \multicolumn{1}{c|}{{\color[HTML]{000000} 99.61}}          & \multicolumn{1}{c|}{{\color[HTML]{000000} 99.69}}          & \multicolumn{1}{c|}{{\color[HTML]{000000} \textbf{99.70}}} & \multicolumn{1}{c|}{{\color[HTML]{000000} 99.67}}          & \multicolumn{1}{c|}{{\color[HTML]{000000} \textbf{99.71}}} & {\color[HTML]{000000} 99.71}          \\ \cline{3-12} 
\multicolumn{2}{|c|}{\multirow{-4}{*}{{\color[HTML]{000000} \begin{tabular}[c]{@{}c@{}}ID class: \\ classification\\  accuracy\end{tabular}}}} & {\color[HTML]{000000} \textbf{Ours}}                           & \multicolumn{1}{c|}{{\color[HTML]{000000} 99.34}}          & \multicolumn{1}{c|}{{\color[HTML]{000000} 99.34}}          & {\color[HTML]{000000} 99.34}          & \multicolumn{1}{c|}{{\color[HTML]{000000} 99.66}}          & \multicolumn{1}{c|}{{\color[HTML]{000000} 99.68}}          & \multicolumn{1}{c|}{{\color[HTML]{000000} 99.68}}          & \multicolumn{1}{c|}{{\color[HTML]{000000} 99.64}}          & \multicolumn{1}{c|}{{\color[HTML]{000000} 99.70}}          & {\color[HTML]{000000} \textbf{99.72}} \\ \hline \hline
\multicolumn{2}{|c|}{{\color[HTML]{000000} }}                                                                                                  & {\color[HTML]{000000}  \cite{softmax}}                                 & \multicolumn{1}{c|}{{\color[HTML]{000000} 43.93}}          & \multicolumn{1}{c|}{{\color[HTML]{000000} 47.20}}          & {\color[HTML]{000000} 44.99}          & \multicolumn{1}{c|}{{\color[HTML]{000000} 45.59}}          & \multicolumn{1}{c|}{{\color[HTML]{000000} 40.34}}          & \multicolumn{1}{c|}{{\color[HTML]{000000} 40.02}}          & \multicolumn{1}{c|}{{\color[HTML]{000000} 45.16}}          & \multicolumn{1}{c|}{{\color[HTML]{000000} 36.44}}          & {\color[HTML]{000000} 41.76}          \\ \cline{3-12} 
\multicolumn{2}{|c|}{{\color[HTML]{000000} }}                                                                                                  & {\color[HTML]{000000}  \cite{lee2018simple}}                             & \multicolumn{1}{c|}{{\color[HTML]{000000} 50.96}}          & \multicolumn{1}{c|}{{\color[HTML]{000000} 61.69}}          & {\color[HTML]{000000} 52.44}          & \multicolumn{1}{c|}{{\color[HTML]{000000} 88.08}}          & \multicolumn{1}{c|}{{\color[HTML]{000000} 87.80}}          & \multicolumn{1}{c|}{{\color[HTML]{000000} 90.03}}          & \multicolumn{1}{c|}{{\color[HTML]{000000} 92.23}}          & \multicolumn{1}{c|}{{\color[HTML]{000000} 79.06}}          & {\color[HTML]{000000} 91.70}          \\ \cline{3-12} 
\multicolumn{2}{|c|}{{\color[HTML]{000000} }}                                                                                                  & {\color[HTML]{000000}  \cite{deep_mcdd}}                               & \multicolumn{1}{c|}{{\color[HTML]{000000} 83.90}}          & \multicolumn{1}{c|}{{\color[HTML]{000000} 84.36}}          & {\color[HTML]{000000} 84.79}          & \multicolumn{1}{c|}{{\color[HTML]{000000} 93.13}}          & \multicolumn{1}{c|}{{\color[HTML]{000000} 93.51}}          & \multicolumn{1}{c|}{{\color[HTML]{000000} 93.69}}          & \multicolumn{1}{c|}{{\color[HTML]{000000} 93.94}}          & \multicolumn{1}{c|}{{\color[HTML]{000000} 84.34}}          & {\color[HTML]{000000} 84.34}          \\ \cline{3-12} 
\multicolumn{2}{|c|}{\multirow{-4}{*}{{\color[HTML]{000000} \begin{tabular}[c]{@{}c@{}}OOD class:\\  TNR @ 85\%\\  TPR\end{tabular}}}}         & {\color[HTML]{000000} \textbf{Ours}}                           & \multicolumn{1}{c|}{{\color[HTML]{000000} \textbf{86.01}}} & \multicolumn{1}{c|}{{\color[HTML]{000000} \textbf{89.09}}} & {\color[HTML]{000000} \textbf{87.16}} & \multicolumn{1}{c|}{{\color[HTML]{000000} \textbf{97.62}}} & \multicolumn{1}{c|}{{\color[HTML]{000000} \textbf{93.64}}} & \multicolumn{1}{c|}{{\color[HTML]{000000} \textbf{94.15}}} & \multicolumn{1}{c|}{{\color[HTML]{000000} \textbf{94.70}}} & \multicolumn{1}{c|}{{\color[HTML]{000000} \textbf{96.61}}} & {\color[HTML]{000000} \textbf{93.99}} \\ \hline \hline
\multicolumn{2}{|c|}{{\color[HTML]{000000} }}                                                                                                  & {\color[HTML]{000000}  \cite{softmax}}                                 & \multicolumn{1}{c|}{{\color[HTML]{000000} 48.43}}          & \multicolumn{1}{c|}{{\color[HTML]{000000} 45.47}}          & {\color[HTML]{000000} 45.40}          & \multicolumn{1}{c|}{{\color[HTML]{000000} 60.24}}          & \multicolumn{1}{c|}{{\color[HTML]{000000} 57.16}}          & \multicolumn{1}{c|}{{\color[HTML]{000000} 56.13}}          & \multicolumn{1}{c|}{{\color[HTML]{000000} 60.36}}          & \multicolumn{1}{c|}{{\color[HTML]{000000} 57.20}}          & {\color[HTML]{000000} 58.60}          \\ \cline{3-12} 
\multicolumn{2}{|c|}{{\color[HTML]{000000} }}                                                                                                  & {\color[HTML]{000000}  \cite{lee2018simple}}                             & \multicolumn{1}{c|}{{\color[HTML]{000000} 78.09}}          & \multicolumn{1}{c|}{{\color[HTML]{000000} 80.16}}          & {\color[HTML]{000000} 75.10}          & \multicolumn{1}{c|}{{\color[HTML]{000000} 92.75}}          & \multicolumn{1}{c|}{{\color[HTML]{000000} 92.53}}          & \multicolumn{1}{c|}{{\color[HTML]{000000} 93.15}}          & \multicolumn{1}{c|}{{\color[HTML]{000000} 93.66}}          & \multicolumn{1}{c|}{{\color[HTML]{000000} 87.33}}          & {\color[HTML]{000000} 93.18}          \\ \cline{3-12} 
\multicolumn{2}{|c|}{{\color[HTML]{000000} }}                                                                                                  & {\color[HTML]{000000}  \cite{deep_mcdd}}                               & \multicolumn{1}{c|}{{\color[HTML]{000000} 90.65}}          & \multicolumn{1}{c|}{{\color[HTML]{000000} 90.73}}          & {\color[HTML]{000000} 90.29}          & \multicolumn{1}{c|}{{\color[HTML]{000000} 95.88}}          & \multicolumn{1}{c|}{{\color[HTML]{000000} 95.98}}          & \multicolumn{1}{c|}{{\color[HTML]{000000} 95.88}}          & \multicolumn{1}{c|}{{\color[HTML]{000000} 96.00}}          & \multicolumn{1}{c|}{{\color[HTML]{000000} 92.35}}          & {\color[HTML]{000000} 92.35}          \\ \cline{3-12} 
\multicolumn{2}{|c|}{\multirow{-4}{*}{{\color[HTML]{000000} \begin{tabular}[c]{@{}c@{}}OOD class:\\ AUROC\end{tabular}}}}                      & {\color[HTML]{000000} \textbf{Ours}}                           & \multicolumn{1}{c|}{{\color[HTML]{000000} \textbf{91.45}}} & \multicolumn{1}{c|}{{\color[HTML]{000000} \textbf{93.22}}} & {\color[HTML]{000000} \textbf{92.13}} & \multicolumn{1}{c|}{{\color[HTML]{000000} \textbf{97.59}}} & \multicolumn{1}{c|}{{\color[HTML]{000000} \textbf{96.04}}} & \multicolumn{1}{c|}{{\color[HTML]{000000} \textbf{96.07}}} & \multicolumn{1}{c|}{{\color[HTML]{000000} \textbf{96.62}}} & \multicolumn{1}{c|}{{\color[HTML]{000000} \textbf{97.11}}} & {\color[HTML]{000000} \textbf{96.12}} \\ \hline \hline
\multicolumn{2}{|c|}{{\color[HTML]{000000} }}                                                                                                  & {\color[HTML]{000000}  \cite{softmax}}                                 & \multicolumn{1}{c|}{{\color[HTML]{000000} 86.12}}          & \multicolumn{1}{c|}{{\color[HTML]{000000} 86.97}}          & {\color[HTML]{000000} 86.83}          & \multicolumn{1}{c|}{{\color[HTML]{000000} 72.44}}          & \multicolumn{1}{c|}{{\color[HTML]{000000} 71.78}}          & \multicolumn{1}{c|}{{\color[HTML]{000000} 70.86}}          & \multicolumn{1}{c|}{{\color[HTML]{000000} 74.61}}          & \multicolumn{1}{c|}{{\color[HTML]{000000} 70.18}}          & {\color[HTML]{000000} 72.06}          \\ \cline{3-12} 
\multicolumn{2}{|c|}{{\color[HTML]{000000} }}                                                                                                  & {\color[HTML]{000000}  \cite{lee2018simple}}                             & \multicolumn{1}{c|}{{\color[HTML]{000000} 93.57}}          & \multicolumn{1}{c|}{{\color[HTML]{000000} 95.09}}          & {\color[HTML]{000000} 93.55}          & \multicolumn{1}{c|}{{\color[HTML]{000000} 90.70}}          & \multicolumn{1}{c|}{{\color[HTML]{000000} 91.79}}          & \multicolumn{1}{c|}{{\color[HTML]{000000} 93.60}}          & \multicolumn{1}{c|}{{\color[HTML]{000000} 92.97}}          & \multicolumn{1}{c|}{{\color[HTML]{000000} 89.21}}          & {\color[HTML]{000000} 93.63}          \\ \cline{3-12} 
\multicolumn{2}{|c|}{{\color[HTML]{000000} }}                                                                                                  & {\color[HTML]{000000}  \cite{deep_mcdd}}                               & \multicolumn{1}{c|}{{\color[HTML]{000000} 97.39}}          & \multicolumn{1}{c|}{{\color[HTML]{000000} 97.69}}          & {\color[HTML]{000000} 97.70}          & \multicolumn{1}{c|}{{\color[HTML]{000000} 95.43}}          & \multicolumn{1}{c|}{{\color[HTML]{000000} \textbf{96.29}}} & \multicolumn{1}{c|}{{\color[HTML]{000000} 95.85}}          & \multicolumn{1}{c|}{{\color[HTML]{000000} 96.09}}          & \multicolumn{1}{c|}{{\color[HTML]{000000} 92.64}}          & {\color[HTML]{000000} 92.64}          \\ \cline{3-12} 
\multicolumn{2}{|c|}{\multirow{-4}{*}{{\color[HTML]{000000} \begin{tabular}[c]{@{}c@{}}OOD class:\\ AUPR\end{tabular}}}}                       & {\color[HTML]{000000} \textbf{Ours}}                           & \multicolumn{1}{c|}{{\color[HTML]{000000} \textbf{97.53}}} & \multicolumn{1}{c|}{{\color[HTML]{000000} \textbf{98.34}}} & {\color[HTML]{000000} \textbf{98.00}} & \multicolumn{1}{c|}{{\color[HTML]{000000} \textbf{97.19}}} & \multicolumn{1}{c|}{{\color[HTML]{000000} 95.98}}          & \multicolumn{1}{c|}{{\color[HTML]{000000} \textbf{96.21}}} & \multicolumn{1}{c|}{{\color[HTML]{000000} \textbf{96.51}}} & \multicolumn{1}{c|}{{\color[HTML]{000000} \textbf{97.04}}} & {\color[HTML]{000000} \textbf{96.28}} \\ \hline \hline
\multicolumn{2}{|c|}{{\color[HTML]{000000} }}                                                                                                  & {\color[HTML]{000000}  \cite{softmax}}                                 & \multicolumn{1}{c|}{{\color[HTML]{000000} 98.59}}          & \multicolumn{1}{c|}{{\color[HTML]{000000} 95.87}}          & {\color[HTML]{000000} 94.47}          & \multicolumn{1}{c|}{{\color[HTML]{000000} 99.61}}          & \multicolumn{1}{c|}{{\color[HTML]{000000} 98.05}}          & \multicolumn{1}{c|}{{\color[HTML]{000000} 97.08}}          & \multicolumn{1}{c|}{{\color[HTML]{000000} 97.44}}          & \multicolumn{1}{c|}{{\color[HTML]{000000} 96.71}}          & {\color[HTML]{000000} 96.00}          \\ \cline{3-12} 
\multicolumn{2}{|c|}{{\color[HTML]{000000} }}                                                                                                  & {\color[HTML]{000000}  \cite{lee2018simple}}                             & \multicolumn{1}{c|}{{\color[HTML]{000000} 15.80}}          & \multicolumn{1}{c|}{{\color[HTML]{000000} 04.44}}           & {\color[HTML]{000000} 04.95}           & \multicolumn{1}{c|}{{\color[HTML]{000000} 24.81}}          & \multicolumn{1}{c|}{{\color[HTML]{000000} 06.32}}           & \multicolumn{1}{c|}{{\color[HTML]{000000} 05.27}}           & \multicolumn{1}{c|}{{\color[HTML]{000000} 04.36}}           & \multicolumn{1}{c|}{{\color[HTML]{000000} 03.04}}           & {\color[HTML]{000000} 01.91}           \\ \cline{3-12} 
\multicolumn{2}{|c|}{{\color[HTML]{000000} }}                                                                                                  & {\color[HTML]{000000}  \cite{deep_mcdd}}                               & \multicolumn{1}{c|}{{\color[HTML]{000000} 99.31}}          & \multicolumn{1}{c|}{{\color[HTML]{000000} 96.33}}          & {\color[HTML]{000000} 94.97}          & \multicolumn{1}{c|}{{\color[HTML]{000000} \textbf{99.63}}} & \multicolumn{1}{c|}{{\color[HTML]{000000} 97.69}}          & \multicolumn{1}{c|}{{\color[HTML]{000000} \textbf{98.05}}} & \multicolumn{1}{c|}{{\color[HTML]{000000} \textbf{97.59}}} & \multicolumn{1}{c|}{{\color[HTML]{000000} 96.23}}          & {\color[HTML]{000000} \textbf{96.23}} \\ \cline{3-12} 
\multicolumn{2}{|c|}{\multirow{-4}{*}{{\color[HTML]{000000} \begin{tabular}[c]{@{}c@{}}Minority class:\\ AUPR\end{tabular}}}}                  & {\color[HTML]{000000} \textbf{Ours}}                           & \multicolumn{1}{c|}{{\color[HTML]{000000} \textbf{99.33}}} & \multicolumn{1}{c|}{{\color[HTML]{000000} \textbf{96.56}}} & {\color[HTML]{000000} \textbf{95.66}} & \multicolumn{1}{c|}{{\color[HTML]{000000} 99.60}}          & \multicolumn{1}{c|}{{\color[HTML]{000000} \textbf{98.05}}} & \multicolumn{1}{c|}{{\color[HTML]{000000} 97.95}}          & \multicolumn{1}{c|}{{\color[HTML]{000000} 97.35}}          & \multicolumn{1}{c|}{{\color[HTML]{000000} \textbf{96.76}}} & {\color[HTML]{000000} \textbf{96.23}} \\ \hline
\end{tabular}
        }
        \caption{Imbalanced dataset: Synthetic financial and Gas Sensor Dataset ~\cite{data_gas}}
        \label{subtab:mnist}
    \end{subtable}
    \\
    \begin{subtable}{\textwidth}
        \centering
        \resizebox{12.45 cm}{!}
       { 
\begin{tabular}{|clc|cccccc|cccccc|}
\hline
\multicolumn{3}{|c|}{{\color[HTML]{000000} Dataset}}                                                                                                                                                                          & \multicolumn{6}{c|}{{\color[HTML]{000000} \begin{tabular}[c]{@{}c@{}}Drive Diagnosis\\ Dataset\end{tabular}}}                                                                                                                                                                                                                                          & \multicolumn{6}{c|}{{\color[HTML]{000000} \begin{tabular}[c]{@{}c@{}}MNIST\\ Dataset\end{tabular}}}                                                                                                                                                                                                                                                    \\ \hline
\multicolumn{3}{|c|}{{\color[HTML]{000000} \begin{tabular}[c]{@{}c@{}}Minority-class \\Down-Sampling \\ Ratio (MDSR)\end{tabular}}}                                                                                                     & \multicolumn{1}{c|}{{\color[HTML]{000000} 1}}              & \multicolumn{1}{c|}{{\color[HTML]{000000} 0.3}}            & \multicolumn{1}{c|}{{\color[HTML]{000000} 0.25}}           & \multicolumn{1}{c|}{{\color[HTML]{000000} 0.2}}            & \multicolumn{1}{c|}{{\color[HTML]{000000} 0.15}}           & {\color[HTML]{000000} 0.1}            & \multicolumn{1}{c|}{{\color[HTML]{000000} 1}}              & \multicolumn{1}{c|}{{\color[HTML]{000000} 0.3}}            & \multicolumn{1}{c|}{{\color[HTML]{000000} 0.25}}           & \multicolumn{1}{c|}{{\color[HTML]{000000} 0.2}}            & \multicolumn{1}{c|}{{\color[HTML]{000000} 0.15}}           & {\color[HTML]{000000} 0.1}            \\ \hline \hline
\multicolumn{2}{|c|}{{\color[HTML]{000000} }}                                                                                                & {\color[HTML]{000000} \cite{softmax}}                         & \multicolumn{1}{c|}{{\color[HTML]{000000} 99.70}}          & \multicolumn{1}{c|}{{\color[HTML]{000000} 99.73}}          & \multicolumn{1}{c|}{{\color[HTML]{000000} 99.70}}          & \multicolumn{1}{c|}{{\color[HTML]{000000} 99.73}}          & \multicolumn{1}{c|}{{\color[HTML]{000000} 99.70}}          & {\color[HTML]{000000} 99.71}          & \multicolumn{1}{c|}{{\color[HTML]{000000} \textbf{97.81}}} & \multicolumn{1}{c|}{{\color[HTML]{000000} \textbf{97.69}}} & \multicolumn{1}{c|}{{\color[HTML]{000000} \textbf{97.68}}} & \multicolumn{1}{c|}{{\color[HTML]{000000} \textbf{97.74}}} & \multicolumn{1}{c|}{{\color[HTML]{000000} \textbf{97.73}}} & {\color[HTML]{000000} \textbf{97.63}} \\ \cline{3-15} 
\multicolumn{2}{|c|}{{\color[HTML]{000000} }}                                                                                                & {\color[HTML]{000000} \cite{lee2018simple}}                   & \multicolumn{1}{c|}{{\color[HTML]{000000} 99.76}}          & \multicolumn{1}{c|}{{\color[HTML]{000000} 94.73}}          & \multicolumn{1}{c|}{{\color[HTML]{000000} 94.77}}          & \multicolumn{1}{c|}{{\color[HTML]{000000} 95.09}}          & \multicolumn{1}{c|}{{\color[HTML]{000000} 94.89}}          & {\color[HTML]{000000} 93.90}          & \multicolumn{1}{c|}{{\color[HTML]{000000} 96.41}}          & \multicolumn{1}{c|}{{\color[HTML]{000000} 96.07}}          & \multicolumn{1}{c|}{{\color[HTML]{000000} 97.63}}          & \multicolumn{1}{c|}{{\color[HTML]{000000} 95.99}}          & \multicolumn{1}{c|}{{\color[HTML]{000000} 96.01}}          & {\color[HTML]{000000} 95.92}          \\ \cline{3-15} 
\multicolumn{2}{|c|}{{\color[HTML]{000000} }}                                                                                                & {\color[HTML]{000000} \cite{deep_mcdd}}                      & \multicolumn{1}{c|}{{\color[HTML]{000000} 95.22}}          & \multicolumn{1}{c|}{{\color[HTML]{000000} 99.79}}          & \multicolumn{1}{c|}{{\color[HTML]{000000} 99.78}}          & \multicolumn{1}{c|}{{\color[HTML]{000000} 99.79}}          & \multicolumn{1}{c|}{{\color[HTML]{000000} 99.74}}          & {\color[HTML]{000000} 99.76}          & \multicolumn{1}{c|}{{\color[HTML]{000000} 97.13}}          & \multicolumn{1}{c|}{{\color[HTML]{000000} 97.19}}          & \multicolumn{1}{c|}{{\color[HTML]{000000} 97.16}}          & \multicolumn{1}{c|}{{\color[HTML]{000000} 97.28}}          & \multicolumn{1}{c|}{{\color[HTML]{000000} 97.05}}          & {\color[HTML]{000000} 97.20}          \\ \cline{3-15} 
\multicolumn{2}{|c|}{\multirow{-4}{*}{{\color[HTML]{000000} \begin{tabular}[c]{@{}c@{}}ID :\\ classi-\\ fication \\ accuracy\end{tabular}}}} & {\color[HTML]{000000} \textbf{Ours}}                                           & \multicolumn{1}{c|}{{\color[HTML]{000000} \textbf{99.81}}} & \multicolumn{1}{c|}{{\color[HTML]{000000} \textbf{99.82}}} & \multicolumn{1}{c|}{{\color[HTML]{000000} \textbf{99.79}}} & \multicolumn{1}{c|}{{\color[HTML]{000000} \textbf{99.81}}} & \multicolumn{1}{c|}{{\color[HTML]{000000} \textbf{99.76}}} & {\color[HTML]{000000} \textbf{99.79}} & \multicolumn{1}{c|}{{\color[HTML]{000000} 97.31}}          & \multicolumn{1}{c|}{{\color[HTML]{000000} 97.09}}          & \multicolumn{1}{c|}{{\color[HTML]{000000} 97.18}}          & \multicolumn{1}{c|}{{\color[HTML]{000000} 97.22}}          & \multicolumn{1}{c|}{{\color[HTML]{000000} 97.19}}          & {\color[HTML]{000000} 97.16}          \\ \hline \hline
\multicolumn{2}{|c|}{{\color[HTML]{000000} }}                                                                                                & {\color[HTML]{000000} \cite{softmax}}                         & \multicolumn{1}{c|}{{\color[HTML]{000000} 15.45}}          & \multicolumn{1}{c|}{{\color[HTML]{000000} 15.68}}          & \multicolumn{1}{c|}{{\color[HTML]{000000} 16.19}}          & \multicolumn{1}{c|}{{\color[HTML]{000000} 17.79}}          & \multicolumn{1}{c|}{{\color[HTML]{000000} 16.89}}          & {\color[HTML]{000000} 19.86}          & \multicolumn{1}{c|}{{\color[HTML]{000000} 82.79}}          & \multicolumn{1}{c|}{{\color[HTML]{000000} 80.51}}          & \multicolumn{1}{c|}{{\color[HTML]{000000} 83.43}}          & \multicolumn{1}{c|}{{\color[HTML]{000000} 80.81}}          & \multicolumn{1}{c|}{{\color[HTML]{000000} 84.85}}          & {\color[HTML]{000000} 86.53}          \\ \cline{3-15} 
\multicolumn{2}{|c|}{{\color[HTML]{000000} }}                                                                                                & {\color[HTML]{000000} \cite{lee2018simple}}                   & \multicolumn{1}{c|}{{\color[HTML]{000000} 41.95}}          & \multicolumn{1}{c|}{{\color[HTML]{000000} 38.30}}          & \multicolumn{1}{c|}{{\color[HTML]{000000} 32.99}}          & \multicolumn{1}{c|}{{\color[HTML]{000000} 43.76}}          & \multicolumn{1}{c|}{{\color[HTML]{000000} 34.72}}          & {\color[HTML]{000000} 35.27}          & \multicolumn{1}{c|}{{\color[HTML]{000000} 59.08}}          & \multicolumn{1}{c|}{{\color[HTML]{000000} 54.99}}          & \multicolumn{1}{c|}{{\color[HTML]{000000} 46.49}}          & \multicolumn{1}{c|}{{\color[HTML]{000000} 54.23}}          & \multicolumn{1}{c|}{{\color[HTML]{000000} 55.02}}          & {\color[HTML]{000000} 51.44}          \\ \cline{3-15} 
\multicolumn{2}{|c|}{{\color[HTML]{000000} }}                                                                                                & {\color[HTML]{000000} \cite{deep_mcdd}}                      & \multicolumn{1}{c|}{{\color[HTML]{000000} 63.56}}          & \multicolumn{1}{c|}{{\color[HTML]{000000} 60.33}}          & \multicolumn{1}{c|}{{\color[HTML]{000000} 53.79}}          & \multicolumn{1}{c|}{{\color[HTML]{000000} 66.61}}          & \multicolumn{1}{c|}{{\color[HTML]{000000} 67.35}}          & {\color[HTML]{000000} 58.79}          & \multicolumn{1}{c|}{{\color[HTML]{000000} 86.33}}          & \multicolumn{1}{c|}{{\color[HTML]{000000} 87.35}}          & \multicolumn{1}{c|}{{\color[HTML]{000000} 82.16}}          & \multicolumn{1}{c|}{{\color[HTML]{000000} 87.46}}          & \multicolumn{1}{c|}{{\color[HTML]{000000} 82.50}}          & {\color[HTML]{000000} 86.25}          \\ \cline{3-15} 
\multicolumn{2}{|c|}{\multirow{-4}{*}{{\color[HTML]{000000} \begin{tabular}[c]{@{}c@{}}OOD: \\ TNR \\ @85\% \\ TPR\end{tabular}}}}          & {\color[HTML]{000000} \textbf{Ours}}                                           & \multicolumn{1}{c|}{{\color[HTML]{000000} \textbf{66.54}}} & \multicolumn{1}{c|}{{\color[HTML]{000000} \textbf{70.08}}} & \multicolumn{1}{c|}{{\color[HTML]{000000} \textbf{64.90}}} & \multicolumn{1}{c|}{{\color[HTML]{000000} \textbf{70.20}}} & \multicolumn{1}{c|}{{\color[HTML]{000000} \textbf{78.04}}} & {\color[HTML]{000000} \textbf{66.13}} & \multicolumn{1}{c|}{{\color[HTML]{000000} \textbf{89.15}}} & \multicolumn{1}{c|}{{\color[HTML]{000000} \textbf{89.46}}} & \multicolumn{1}{c|}{{\color[HTML]{000000} \textbf{89.12}}} & \multicolumn{1}{c|}{{\color[HTML]{000000} \textbf{89.93}}} & \multicolumn{1}{c|}{{\color[HTML]{000000} \textbf{87.72}}} & {\color[HTML]{000000} \textbf{88.77}} \\ \hline \hline
\multicolumn{2}{|c|}{{\color[HTML]{000000} }}                                                                                                & {\color[HTML]{000000} \cite{softmax}}                         & \multicolumn{1}{c|}{{\color[HTML]{000000} 23.52}}          & \multicolumn{1}{c|}{{\color[HTML]{000000} 22.50}}          & \multicolumn{1}{c|}{{\color[HTML]{000000} 22.39}}          & \multicolumn{1}{c|}{{\color[HTML]{000000} 26.48}}          & \multicolumn{1}{c|}{{\color[HTML]{000000} 24.77}}          & {\color[HTML]{000000} 26.20}          & \multicolumn{1}{c|}{{\color[HTML]{000000} 80.14}}          & \multicolumn{1}{c|}{{\color[HTML]{000000} 78.05}}          & \multicolumn{1}{c|}{{\color[HTML]{000000} 80.73}}          & \multicolumn{1}{c|}{{\color[HTML]{000000} 78.16}}          & \multicolumn{1}{c|}{{\color[HTML]{000000} 82.23}}          & {\color[HTML]{000000} 84.12}          \\ \cline{3-15} 
\multicolumn{2}{|c|}{{\color[HTML]{000000} }}                                                                                                & {\color[HTML]{000000} \cite{lee2018simple}}                   & \multicolumn{1}{c|}{{\color[HTML]{000000} 78.87}}          & \multicolumn{1}{c|}{{\color[HTML]{000000} 79.37}}          & \multicolumn{1}{c|}{{\color[HTML]{000000} 75.22}}          & \multicolumn{1}{c|}{{\color[HTML]{000000} 77.95}}          & \multicolumn{1}{c|}{{\color[HTML]{000000} 74.84}}          & {\color[HTML]{000000} 70.99}          & \multicolumn{1}{c|}{{\color[HTML]{000000} 81.07}}          & \multicolumn{1}{c|}{{\color[HTML]{000000} 78.16}}          & \multicolumn{1}{c|}{{\color[HTML]{000000} 75.07}}          & \multicolumn{1}{c|}{{\color[HTML]{000000} 77.48}}          & \multicolumn{1}{c|}{{\color[HTML]{000000} 78.15}}          & {\color[HTML]{000000} 76.54}          \\ \cline{3-15} 
\multicolumn{2}{|c|}{{\color[HTML]{000000} }}                                                                                                & {\color[HTML]{000000} \cite{deep_mcdd}}                      & \multicolumn{1}{c|}{{\color[HTML]{000000} 80.29}}          & \multicolumn{1}{c|}{{\color[HTML]{000000} 77.02}}          & \multicolumn{1}{c|}{{\color[HTML]{000000} 75.31}}          & \multicolumn{1}{c|}{{\color[HTML]{000000} 82.97}}          & \multicolumn{1}{c|}{{\color[HTML]{000000} 82.58}}          & {\color[HTML]{000000} 78.07}          & \multicolumn{1}{c|}{{\color[HTML]{000000} 91.36}}          & \multicolumn{1}{c|}{{\color[HTML]{000000} 92.13}}          & \multicolumn{1}{c|}{{\color[HTML]{000000} 89.87}}          & \multicolumn{1}{c|}{{\color[HTML]{000000} 86.48}}          & \multicolumn{1}{c|}{{\color[HTML]{000000} 89.41}}          & {\color[HTML]{000000} 91.56}          \\ \cline{3-15} 
\multicolumn{2}{|c|}{\multirow{-4}{*}{{\color[HTML]{000000} \begin{tabular}[c]{@{}c@{}}OOD:\\ AUROC\end{tabular}}}}                         & {\color[HTML]{000000} \textbf{Ours}}                                           & \multicolumn{1}{c|}{{\color[HTML]{000000} \textbf{83.43}}} & \multicolumn{1}{c|}{{\color[HTML]{000000} \textbf{84.94}}} & \multicolumn{1}{c|}{{\color[HTML]{000000} \textbf{82.85}}} & \multicolumn{1}{c|}{{\color[HTML]{000000} \textbf{85.44}}} & \multicolumn{1}{c|}{{\color[HTML]{000000} \textbf{88.74}}} & {\color[HTML]{000000} \textbf{83.59}} & \multicolumn{1}{c|}{{\color[HTML]{000000} \textbf{93.16}}} & \multicolumn{1}{c|}{{\color[HTML]{000000} \textbf{93.22}}} & \multicolumn{1}{c|}{{\color[HTML]{000000} \textbf{92.97}}} & \multicolumn{1}{c|}{{\color[HTML]{000000} \textbf{93.49}}} & \multicolumn{1}{c|}{{\color[HTML]{000000} \textbf{92.52}}} & {\color[HTML]{000000} \textbf{92.81}} \\ \hline \hline
\multicolumn{2}{|c|}{{\color[HTML]{000000} }}                                                                                                & {\color[HTML]{000000} \cite{softmax}}                         & \multicolumn{1}{c|}{{\color[HTML]{000000} 28.82}}          & \multicolumn{1}{c|}{{\color[HTML]{000000} 29.71}}          & \multicolumn{1}{c|}{{\color[HTML]{000000} 30.02}}          & \multicolumn{1}{c|}{{\color[HTML]{000000} 32.51}}          & \multicolumn{1}{c|}{{\color[HTML]{000000} 31.10}}          & {\color[HTML]{000000} 33.32}          & \multicolumn{1}{c|}{{\color[HTML]{000000} 79.55}}          & \multicolumn{1}{c|}{{\color[HTML]{000000} 79.39}}          & \multicolumn{1}{c|}{{\color[HTML]{000000} 80.98}}          & \multicolumn{1}{c|}{{\color[HTML]{000000} 79.27}}          & \multicolumn{1}{c|}{{\color[HTML]{000000} 82.51}}          & {\color[HTML]{000000} 83.16}          \\ \cline{3-15} 
\multicolumn{2}{|c|}{{\color[HTML]{000000} }}                                                                                                & {\color[HTML]{000000} \cite{lee2018simple}}                   & \multicolumn{1}{c|}{{\color[HTML]{000000} 54.48}}          & \multicolumn{1}{c|}{{\color[HTML]{000000} 54.33}}          & \multicolumn{1}{c|}{{\color[HTML]{000000} 52.18}}          & \multicolumn{1}{c|}{{\color[HTML]{000000} 57.47}}          & \multicolumn{1}{c|}{{\color[HTML]{000000} 53.35}}          & {\color[HTML]{000000} 52.29}          & \multicolumn{1}{c|}{{\color[HTML]{000000} 67.15}}          & \multicolumn{1}{c|}{{\color[HTML]{000000} 64.83}}          & \multicolumn{1}{c|}{{\color[HTML]{000000} 61.01}}          & \multicolumn{1}{c|}{{\color[HTML]{000000} 65.65}}          & \multicolumn{1}{c|}{{\color[HTML]{000000} 66.59}}          & {\color[HTML]{000000} 63.44}          \\ \cline{3-15} 
\multicolumn{2}{|c|}{{\color[HTML]{000000} }}                                                                                                & {\color[HTML]{000000} \cite{deep_mcdd}}                      & \multicolumn{1}{c|}{{\color[HTML]{000000} 70.31}}          & \multicolumn{1}{c|}{{\color[HTML]{000000} 69.98}}          & \multicolumn{1}{c|}{{\color[HTML]{000000} 66.13}}          & \multicolumn{1}{c|}{{\color[HTML]{000000} 74.30}}          & \multicolumn{1}{c|}{{\color[HTML]{000000} 74.77}}          & {\color[HTML]{000000} 69.50}          & \multicolumn{1}{c|}{{\color[HTML]{000000} 83.94}}          & \multicolumn{1}{c|}{{\color[HTML]{000000} 86.95}}          & \multicolumn{1}{c|}{{\color[HTML]{000000} 83.63}}          & \multicolumn{1}{c|}{{\color[HTML]{000000} 87.70}}          & \multicolumn{1}{c|}{{\color[HTML]{000000} 83.80}}          & {\color[HTML]{000000} 86.47}          \\ \cline{3-15} 
\multicolumn{2}{|c|}{\multirow{-4}{*}{{\color[HTML]{000000} \begin{tabular}[c]{@{}c@{}}OOD:\\ AUPR\end{tabular}}}}                          & {\color[HTML]{000000} \textbf{Ours}}                                           & \multicolumn{1}{c|}{{\color[HTML]{000000} \textbf{71.44}}} & \multicolumn{1}{c|}{{\color[HTML]{000000} \textbf{74.47}}} & \multicolumn{1}{c|}{{\color[HTML]{000000} \textbf{71.94}}} & \multicolumn{1}{c|}{{\color[HTML]{000000} \textbf{75.76}}} & \multicolumn{1}{c|}{{\color[HTML]{000000} \textbf{80.42}}} & {\color[HTML]{000000} \textbf{75.30}} & \multicolumn{1}{c|}{{\color[HTML]{000000} \textbf{87.64}}} & \multicolumn{1}{c|}{{\color[HTML]{000000} \textbf{88.52}}} & \multicolumn{1}{c|}{{\color[HTML]{000000} \textbf{88.08}}} & \multicolumn{1}{c|}{{\color[HTML]{000000} \textbf{89.30}}} & \multicolumn{1}{c|}{{\color[HTML]{000000} \textbf{88.41}}} & {\color[HTML]{000000} \textbf{88.58}} \\ \hline \hline
\multicolumn{2}{|c|}{{\color[HTML]{000000} }}                                                                                                & {\color[HTML]{000000} \cite{softmax}}                         & \multicolumn{1}{c|}{{\color[HTML]{000000} 99.71}}          & \multicolumn{1}{c|}{{\color[HTML]{000000} 99.48}}          & \multicolumn{1}{c|}{{\color[HTML]{000000} 98.96}}          & \multicolumn{1}{c|}{{\color[HTML]{000000} 98.75}}          & \multicolumn{1}{c|}{{\color[HTML]{000000} 99.10}}          & {\color[HTML]{000000} 98.58}          & \multicolumn{1}{c|}{{\color[HTML]{000000} 99.21}}          & \multicolumn{1}{c|}{{\color[HTML]{000000} 98.98}}          & \multicolumn{1}{c|}{{\color[HTML]{000000} \textbf{98.91}}} & \multicolumn{1}{c|}{{\color[HTML]{000000} 98.10}}          & \multicolumn{1}{c|}{{\color[HTML]{000000} 98.06}}          & {\color[HTML]{000000} \textbf{97.33}} \\ \cline{3-15} 
\multicolumn{2}{|c|}{{\color[HTML]{000000} }}                                                                                                & {\color[HTML]{000000} \cite{lee2018simple}}                   & \multicolumn{1}{c|}{{\color[HTML]{000000} 21.90}}          & \multicolumn{1}{c|}{{\color[HTML]{000000} 04.83}}           & \multicolumn{1}{c|}{{\color[HTML]{000000} 04.67}}           & \multicolumn{1}{c|}{{\color[HTML]{000000} 04.53}}           & \multicolumn{1}{c|}{{\color[HTML]{000000} 03.02}}           & {\color[HTML]{000000} 01.61}           & \multicolumn{1}{c|}{{\color[HTML]{000000} 80.72}}          & \multicolumn{1}{c|}{{\color[HTML]{000000} 58.04}}          & \multicolumn{1}{c|}{{\color[HTML]{000000} 56.45}}          & \multicolumn{1}{c|}{{\color[HTML]{000000} 55.48}}          & \multicolumn{1}{c|}{{\color[HTML]{000000} 53.82}}          & {\color[HTML]{000000} 41.66}          \\ \cline{3-15} 
\multicolumn{2}{|c|}{{\color[HTML]{000000} }}                                                                                                & {\color[HTML]{000000} \cite{deep_mcdd}}                      & \multicolumn{1}{c|}{{\color[HTML]{000000} 99.83}}          & \multicolumn{1}{c|}{{\color[HTML]{000000} 99.43}}          & \multicolumn{1}{c|}{{\color[HTML]{000000} 99.47}}          & \multicolumn{1}{c|}{{\color[HTML]{000000} \textbf{99.33}}} & \multicolumn{1}{c|}{{\color[HTML]{000000} 98.70}}          & {\color[HTML]{000000} 98.13}          & \multicolumn{1}{c|}{{\color[HTML]{000000} 99.64}}          & \multicolumn{1}{c|}{{\color[HTML]{000000} \textbf{99.10}}} & \multicolumn{1}{c|}{{\color[HTML]{000000} 98.90}}          & \multicolumn{1}{c|}{{\color[HTML]{000000} 98.50}}          & \multicolumn{1}{c|}{{\color[HTML]{000000} 98.15}}          & {\color[HTML]{000000} 97.05}          \\ \cline{3-15} 
\multicolumn{2}{|c|}{\multirow{-4}{*}{{\color[HTML]{000000} \begin{tabular}[c]{@{}c@{}}Minority:\\ AUPR\end{tabular}}}}                     & {\color[HTML]{000000} \textbf{Ours}}                                           & \multicolumn{1}{c|}{{\color[HTML]{000000} \textbf{99.85}}} & \multicolumn{1}{c|}{{\color[HTML]{000000} \textbf{99.53}}} & \multicolumn{1}{c|}{{\color[HTML]{000000} \textbf{99.48}}} & \multicolumn{1}{c|}{{\color[HTML]{000000} 99.32}}          & \multicolumn{1}{c|}{{\color[HTML]{000000} \textbf{99.21}}} & {\color[HTML]{000000} \textbf{98.83}} & \multicolumn{1}{c|}{{\color[HTML]{000000} \textbf{99.72}}} & \multicolumn{1}{c|}{{\color[HTML]{000000} 98.94}}          & \multicolumn{1}{c|}{{\color[HTML]{000000} 98.81}}          & \multicolumn{1}{c|}{{\color[HTML]{000000} \textbf{98.56}}} & \multicolumn{1}{c|}{{\color[HTML]{000000} \textbf{98.19}}} & {\color[HTML]{000000} 96.81}          \\ \hline
\end{tabular}
}
        \caption{Balanced dataset: Drive Diagnosis \cite{data_drive} and MNIST Dataset ~\cite{data_mnist}}
        \label{subtab:gas}
    \end{subtable}
    
     \caption{Comparison of proposed DNN-GDITD with Softmax \cite{softmax}, Mahalanobis \cite{lee2018simple} and Deep-MCDD \cite{deep_mcdd} (Best in bold).}
    \label{table_class}
\end{table}

Moreover, we observe that as the imbalance in data increases (i.e., as MDSR decreases), the OOD score drastically decreases for all the baselines. However, the proposed DNN-GDITD (ours) performs consistently at different MDSR values for OOD detection with an average boost of $3.32\%$ compared to all the baselines. Subsequently, to ascertain whether the DNN-GDITD and second-best scores are statistically different, we conducted a Wilcoxon signed-rank test for paired samples at a significance level of $0.05$. For detecting the OOD class, we obtained p-values of $9.5\times10^{-7}$, $9.5\times10^{-7}$, and $4.5\times10^{-6}$ for TNR, AUROC, and AUPR, respectively, indicating statistical significance. Further, we also observe that using spherical decision boundaries (Deep-MCDD and proposed DNN-GDITD) enhances OOD detection compared to linear decision boundaries (Softmax, Mahalanobis).

\subsection{ Ablation}
We conducted an ablation study to demonstrate the effectiveness of each loss component proposed in Table \ref{ablation_table}. In the DNN-GDITD algorithm, we have 4 losses functions i.e., pull loss (\ref{pull_loss}), SL (\ref{sl_loss}), $EFL_1$ (\ref{efl1_loss}) and $EFL_2$ (\ref{efl2_loss}). We show the the losses performance on the Gas sensor \cite{data_gas} dataset when only one out of 4 loss is considered vs when only one loss is omitted using the metrics mentioned in Section \ref{evaluation}. Table \ref{ablation_table} suggests that performance without an individual loss (the remaining other three losses used in training) is greater than the  performance obtained with the individual loss only. The only exception is when only the score-based loss is employed. The score-based loss ensures that the score value $\zeta_i(x)$ is non-negative from its  class and negative for the rest of the classes, thereby enforcing that the data point should belong to its own cluster. However, comparing the AUPR score of the OOD class while using only SL with DNN-GDITD, we observe a dip of $2.42\%$ as SL in isolation doesn't ensure compact cluster formation. Further, $EFL_2$ performs poorly for both the OOD and minority classes. As a cross-entropy-based loss, $EFL_2$ tends to increase the difference between scores for the actual class and the rest. However, it does not enforce any condition on the sign of $\zeta_i(x)$. This suggests the need to incorporate SL alongside $EFL_2$ loss. Similarly, $EFL_1$ increases the difference between the distance of a data point for the actual class versus the remaining classes. However, it does not enforce that the distance of the data point from its  class should be close to zero. Thus, a pull loss should be used in conjunction with $EFL_1$. Further, as observed, omitting SL results in poor performance, justifying the necessity of using both distance-based and score-based losses to achieve the best overall performance.

\begin{table}[hbt!]

\centering

\begin{tabular}{|c|ccc|}
\hline
\multirow{2}{*}{\begin{tabular}[c]{@{}c@{}}\textbf{Gas dataset}\\\end{tabular}}  & \multicolumn{3}{c|}{\textit{With only row index loss} \textbf{\textbar} \textit{all but without row index loss}} \\ \cline{2-4} 
& \multicolumn{1}{c|}{\textbf{ID class} : accuracy \;}  & \multicolumn{1}{c|}{\textbf{OOD class}: AUPR \;}                                                                                                             & \textbf{Minority class} : AUPR\;
 \\ \hline\hline
Pull loss                                                                                                      & \multicolumn{1}{c|}{94.72  \textbf{\textbar} 99.63}          & \multicolumn{1}{c|}{77.40 \textbf{\textbar} 96.71}  & 97.19 \textbf{\textbar} 99.51                      \\ \hline
SL                                                                                                      & \multicolumn{1}{c|}{\textbf{99.67} \textbf{\textbar} 99.56}        & \multicolumn{1}{c|}{ 94.77 \textbf{\textbar} 93.75 } & 99.36 \textbf{\textbar}  99.27                     \\ \hline
$EFL_1 $                                                                                                         & \multicolumn{1}{c|}{99.40  \textbf{\textbar} \textbf{99.67}}       & \multicolumn{1}{c|}{97.03 \textbf{\textbar} 95.08}  & 98.58 \textbf{\textbar} 99.57                      \\ \hline
$EFL_2 $                                                                                                         & \multicolumn{1}{c|}{99.57 \textbf{\textbar} 99.62}           & \multicolumn{1}{c|}{43.86 \textbf{\textbar} 97.02} & 13.10 \textbf{\textbar} 99.53                      \\ \hline

\textbf{DNN-GDITD}                                                                                       & \multicolumn{1}{c|}{{99.66}}             & \multicolumn{1}{c|}{{\textbf{97.19}}}     & {\textbf{99.60}}             \\ \hline
\end{tabular}
\vspace{2 pt}
\caption{Ablation: Performance on ID (classes 1,3,4,5,6), OOD (class 0), and minority (class 2) data with \& without each loss on Gas Sensor dataset \cite{data_gas} with MSDR value as 1. Best value for each metric is marked in bold.}
\label{ablation_table}
\end{table}

\section{Significance and Conclusion}\label{conclude}
We propose a novel method (DNN-GDITD) based on Gaussian Discriminator analysis to tackle class imbalance in tabular datasets during training and detect OOD samples while testing. 
DNN-GDITD consists of four loss functions. Pull and EFL1 losses reduce the intra-class distance. SL and EFL2 ensure that the score of a data point for its class is non-negative while the score of the data point from other classes is negative. The \textit{EF1} takes the reciprocal distance of training samples from all the classes to handle the imbalanced setting. In this work, we show experimental evaluation on four benchmark datasets. After comparing DNN-GDITD with current SOTA methods,  we observe that DNN-GDITD gives the best performance in OOD-detection for tabular datasets with an average boost of $3.32\%$ and comparable performance for classifying in-distribution samples. However, as outliers can be present in various domains, we believe DNN-GDITD can be extended to other domains/modalities.

 \newpage
 
\addcontentsline{toc}{section}{References} \label{bib}
\bibliographystyle{IEEEtran}

\begin{thebibliography}{}
\bibitem{finance} Sihem Khemakhem, Fatma Ben Said, Younes Boujelbene (2018). \href{https://doi.org/10.1108/JM2-01-2017-0002}{Credit risk assessment for unbalanced datasets based on data mining, artificial neural network and support vector machines}. Journal of Modelling in Management.

\bibitem{manufac} Jefkine Kafunah, Priyanka Verma, Muhammad Intizar Ali, John G. Breslin (2023). \href{https://doi.org/10.1109/ACCESS.2023.3337658}{Out-of-Distribution Data Generation for Fault Detection and Diagnosis in Industrial Systems}. IEEE Access.

\bibitem{neurips23} Xuefeng Du, Yiyou Sun, Xiaojin Zhu, Yixuan Li (2023). \href{https://arxiv.org/abs/2309.13415}{Dream the Impossible: Outlier Imagination with Diffusion Models}. arXiv.

\bibitem{newsArticle} Jay Ramey (2021). \href{https://www.autoweek.com/news/green-cars/a37114603/tesla-fsd-mistakes-moon-for-traffic-light/}{Tesla FSD Mistakes Moon For Yellow Traffic Light}. [Accessed 28-Dec-2023].

\bibitem{data_mnist} Li Deng (2012). \href{https://doi.org/10.1109/MSP.2012.2211477}{The MNIST Database of Handwritten Digit Images for Machine Learning Research [Best of the Web]}. IEEE Signal Processing Magazine.

\bibitem{deep_mcdd} Dongha Lee, Sehun Yu, Hwanjo Yu (2020). Multi-class data description for out-of-distribution detection. In Proceedings of the 26th ACM SIGKDD International Conference on Knowledge Discovery \& Data Mining.

\bibitem{lee2018simple} Kimin Lee, Kibok Lee, Honglak Lee, Jinwoo Shin (2018). A Simple Unified Framework for Detecting Out-of-Distribution Samples and Adversarial Attacks. Advances in Neural Information Processing Systems 31 (NeurIPS).

\bibitem{data_gas} Alexander Vergara (2012). \href{https://doi.org/10.24432/C5RP6W}{Gas Sensor Array Drift Dataset}. UCI Machine Learning Repository.

\bibitem{adam} Diederik P. Kingma, Jimmy Ba (2015). Adam: A Method for Stochastic Optimization. In Proceedings of the 3rd International Conference on Learning Representations, ICLR 2015, San Diego, CA, USA, May 7-9, 2015, Conference Track Proceedings.

\bibitem{smote} N. V. Chawla, K. W. Bowyer, L. O. Hall, W. P. Kegelmeyer (2002). \href{https://www.jair.org/index.php/jair/article/view/10302}{SMOTE: Synthetic Minority Over-sampling Technique}. Journal of Artificial Intelligence Research.

\bibitem{data_drive} Martyna Bator (2015). \href{https://doi.org/10.24432/C5VP5F}{Dataset for Sensorless Drive Diagnosis}. UCI Machine Learning Repository.

\bibitem{data_shuttle} Creative Commons Attribution 4.0 International Public License (2015). \href{https://doi.org/10.24432/C5WS31}{Statlog (Shuttle)}. UCI Machine Learning Repository.

\bibitem{privacy} Akira Imakura, Masateru Kihira, Yukihiko Okada, Tetsuya Sakurai (2023). \href{https://doi.org/10.1016/j.eswa.2023.120385}{Another Use of SMOTE for Interpretable Data Collaboration Analysis}. Expert Systems with Applications.

\bibitem{ODIN} Shiyu Liang, Yixuan Li, R. Srikant (2018). Enhancing The Reliability of Out-of-distribution Image Detection in Neural Networks. International Conference on Learning Representations (ICLR).

\bibitem{softmax} Dan Hendrycks, Kevin Gimpel (2017). A Baseline for Detecting Misclassified and Out-of-Distribution Examples in Neural Networks. International Conference on Learning Representations.

\bibitem{16} Andrey Malinin, Mark Gales (2018). Predictive Uncertainty Estimation via Prior Networks. Conference on Neural Information Processing Systems (NeurIPS).

\bibitem{17} Seyed-Mohsen Moosavi-Dezfooli, Alhussein Fawzi, Omar Fawzi, Pascal Frossard (2017). Universal adversarial perturbations. Conference on Computer Vision and Pattern Recognition (CVPR).

\bibitem{focal} Tsung-Yi Lin, Priya Goyal, Ross Girshick, Kaiming He, Piotr Dollár (2020). Focal Loss for Dense Object Detection. IEEE Transactions on Pattern Analysis and Machine Intelligence.

\bibitem{effective} Yin Cui, Menglin Jia, Tsung-Yi Lin, Yang Song, Serge Belongie (2019). Class-Balanced Loss Based on Effective Number of Samples. 2019 IEEE/CVF Conference on Computer Vision and Pattern Recognition (CVPR).

\bibitem{26} Christian Szegedy, Wojciech Zaremba, Ilya Sutskever, Joan Bruna, Dumitru Erhan, Ian Goodfellow, Rob Fergus (2014). Intriguing properties of neural networks. International Conference on Learning Representations (ICLR).

\bibitem{bcd} P. Tseng (2001). Convergence of a Block Coordinate Descent Method for Nondifferentiable Minimization. Journal of Optimization Theory and Applications 109.

\bibitem{elflein2021outofdistribution} Sven Elflein, Bertrand Charpentier, Daniel Zügner, Stephan Günnemann (2021). On Out-of-distribution Detection with Energy-based Models. International Conference on Machine Learning(ICML).

\bibitem{guo2017calibration} Chuan Guo, Geoff Pleiss, Yu Sun, Kilian Q. Weinberger (2017). On Calibration of Modern Neural Networks. International Conference on Machine Learning (ICML).

\bibitem{tsne} Laurens van der Maaten, Geoffrey Hinton (2008). \href{http://jmlr.org/papers/v9/vandermaaten08a.html}{Visualizing Data using t-SNE}. Journal of Machine Learning Research (JMLR).

\bibitem{thirdpartyfraud} Clifton Phua, Vincent Cheng-Siong Lee, Kate Smith‐Miles, Ross W. Gayler (2010). \href{https://arxiv.org/abs/1009.6119}{A Comprehensive Survey of Data Mining-based Fraud Detection Research}. ArXiv.

\bibitem{firstpartyfraud} Chioma Vivian Amasiatu, Mahmood Hussain Shah (2018). \href{https://www.emerald.com/insight/content/doi/10.1108/IJRDM-04-2017-0064/full/html}{First party fraud management: framework for the retail industry}. International Journal of Retail \& Distribution Management.

\bibitem{OODmixup} Taocun Yang, Yaping Huang, Yanlin Xie, Junbo Liu, Shengchun Wang (2023). \href{https://doi.org/10.1145/3578935}{MixOOD: Improving Out-of-Distribution Detection with Enhanced Data Mixup}. ACM Trans. Multimedia Comput. Commun. Appl.

\bibitem{Mdnn1} Alex Krizhevsky, Ilya Sutskever, Geoffrey E. Hinton (2012). \href{https://proceedings.neurips.cc/paper/2012/file/c399862d3b9d6b76c8436e924a68c45b-Paper.pdf}{Imagenet classification with deep convolutional neural networks}. Advances in neural information processing systems.

\bibitem{Mdnn2} Karen Simonyan, Andrew Zisserman (2015). Very Deep Convolutional Networks for Large-Scale Image Recognition. In Proceedings of the 3rd International Conference on Learning Representations, ICLR 2015, San Diego, CA, USA, May 7-9, 2015, Conference Track Proceedings. \href{http://arxiv.org/abs/1409.1556}{Link}.

\bibitem{Mdnn3} Kaiming He, Xiangyu Zhang, Shaoqing Ren, Jian Sun (2015). \href{http://arxiv.org/abs/1512.03385}{Deep Residual Learning for Image Recognition}. CoRR.

\bibitem{Mdnn4} C. Chow (1970). \href{https://doi.org/10.1109/TIT.1970.1054406}{On optimum recognition error and reject tradeoff}. IEEE Transactions on Information Theory.

\bibitem{Mdnn5} Chiyuan Zhang, Samy Bengio, Moritz Hardt, Benjamin Recht, Oriol Vinyals (2016). \href{http://arxiv.org/abs/1611.03530}{Understanding deep learning requires rethinking generalization}. CoRR.

\bibitem{softmax1} Khanh Nguyen, Brendan O'Connor (2015). \href{http://arxiv.org/abs/1508.05154}{Posterior calibration and exploratory analysis for natural language processing models}. CoRR.


\bibitem{softmax2} Yuting Zhang, Kibok Lee, Honglak Lee (2016). \href{http://arxiv.org/abs/1606.06582}{Augmenting Supervised Neural Networks with Unsupervised Objectives for Large-scale Image Classification}. CoRR, abs/1606.06582.

\bibitem{softmax3} Foster J. Provost, Tom Fawcett, Ron Kohavi (1998). \href{https://doi.org/10.1016/B978-1-55860-556-3.50067-3}{The Case against Accuracy Estimation for Comparing Induction Algorithms}. Proceedings of the Fifteenth International Conference on Machine Learning, ICML '98, San Francisco, CA, USA, pp. 445–453. Morgan Kaufmann Publishers Inc.

\bibitem{softmax4} Anh Mai Nguyen, Jason Yosinski, Jeff Clune (2014). \href{http://arxiv.org/abs/1412.1897}{Deep Neural Networks are Easily Fooled: High Confidence Predictions for Unrecognizable Images}. CoRR, abs/1412.1897.

\bibitem{sgd} Shun-ichi Amari (1993). \href{https://doi.org/10.1016/0925-2312(93)90010-K}{Backpropagation and stochastic gradient descent method}. Neurocomputing, Vol. 5, No. 4, pp. 185-196.

\bibitem{momen} Ning Qian (1999). \href{https://doi.org/10.1016/S0893-6080(98)00116-6}{On the momentum term in gradient descent learning algorithms}. Neural Networks, Vol. 12, No. 1, pp. 145-151.

\bibitem{ood4} Gabriel Pereyra, George Tucker, Jan Chorowski, Lukasz Kaiser, Geoffrey E. Hinton (2017). \href{http://arxiv.org/abs/1701.06548}{Regularizing Neural Networks by Penalizing Confident Output Distributions}. CoRR, abs/1701.06548.

\bibitem{ood5} Geoffrey E. Hinton, Oriol Vinyals, Jeffrey Dean (2015). \href{https://api.semanticscholar.org/CorpusID:7200347}{Distilling the Knowledge in a Neural Network}. ArXiv, abs/1503.02531.

\bibitem{adv_per1} Ian J. Goodfellow, Jonathon Shlens, Christian Szegedy (2015). \href{http://arxiv.org/abs/1412.6572}{Explaining and Harnessing Adversarial Examples}. 3rd International Conference on Learning Representations, ICLR 2015, San Diego, CA, USA, Conference Track Proceedings.

\bibitem{adv_per2} Dario Amodei, Chris Olah, Jacob Steinhardt, Paul F. Christiano, John Schulman, Dan Mané (2016). \href{http://arxiv.org/abs/1606.06565}{Concrete Problems in AI Safety}. CoRR, abs/1606.06565.


\end{thebibliography}

\end{document}


\sloppy
%

\def\teal{\textcolor{teal}}
\def\red{\textcolor{red}}
\def\green{\textcolor{green}}
\def\hlinewd#1{%
\noalign{\ifnum0=`}\fi\hrule \@height #1 %
\futurelet\reserved@a\@xhline} 
\title{Supplementary material: \\\small{DNN-GDID: Out-of-distribution detection via  Deep Neural Network based Gaussian Descriptor for Imbalanced Data}}
%

\name{Anonymous ICME submission }

\address{}
\maketitle

\section{Additional results} \label{introduction}

We present the pseudo-code for our algorithm, DNN-GDID, in Algorithm \ref{alg:two} to enhance understanding. Additionally, we compare t-SNE (t-distributed Stochastic Neighbor Embedding) plots of Deep-MCDD \cite{deep_mcdd} and our proposed algorithm, DNN-GDID. t-SNE \cite{tsne} is an unsupervised non-linear dimensionality reduction technique used to visualize high-dimensional data. We generate t-SNE plots using the testing data embedding vectors obtained from Deep-MCDD and DNN-GDID when the Gas Sensor dataset \cite{data_gas} is utilized, as shown in Figure \ref{fig3}. We observe that in DNN-GDID, OOD samples are positioned outside the class cluster considerably away when compared to Deep-MCDD's t-SNE plot. This observation suggests that DNN-GDID is capable of creating better spherical decision boundaries by making clusters more compact and distinct.

\begin{figure}[ht] 
\includegraphics[trim={1.7cm  9.8cm 0 1.3cm},clip, width=9.5cm, height= 3.5cm]{DNN-GDID copy.pdf}
\centering
\caption{t-SNE ~\cite{tsne} plots for embeddings obtained from Deep-MCDD ~\cite{deep_mcdd}  vs DNN-GDID (our) algorithm for Gas sensor dataset ~\cite{data_gas}. } \label{fig3}
\end{figure}

\section{Additional experimental details}
The paper \cite{lee2018simple}, utilized as one of the baselines, employs a pre-trained softmax classifier-based model. Subsequently, they introduce a feature ensemble technique that necessitates out-of-distribution (OOD) samples. Since our method does not assume the existence of OOD samples during training in any form, we omit this step and incorporate the results for \cite{lee2018simple} in the result table in the main paper. However, we contend that utilizing the feature ensemble-based Mahalanobis score, or any algorithm employing a softmax-based classifier for OOD
\RestyleAlgo{ruled}

\SetKwComment{Comment}{$\triangleright$ }{}

\begin{algorithm}[hbt!] \footnotesize 
\caption{Pseudo code for DNN-GDID}\label{alg:two}
\KwData{Training (in-distribution) data: $\mathcal{D}_{train} = \{X,Y_X\}$ with $k$ classes, testing data $\mathcal{D}_{test} = \{X',Y_{X'}\}$ with OOD and k classes ( as seen during training), base DNN: $f(:,W) \in \mathbb{R}^d$, mean and standard deviation: $(\mu_i, \sigma_i)$ for $i \in \{1, 2, \ldots, k\}$ for k class clusters in latent space.}
\For{iter  $ \in \{1,2, \dots\}$ }
   {sample a mini-batch $B$ from $\mathcal{D}_{train}$\;
   \For{$x \in B:$}
   { \Comment{\teal{feed sample point into DNN to get vector representation in latent space:}}
   $f(:,W): x \mapsto f(x)$\; 
   \Comment{\teal{Get distance of x from each k class cluster after converting the k-classes distribution in latent space obtained from $f(:,W)$ to isotropic Gaussian distributions:}}
   $D_i(x)= \frac{\norm{f(x) - \mu_i}^2}{2\sigma_i^2} + log(\sigma_i)^d, 1\leq i \leq k$ 
    \Comment{\teal{Predicted class for x:}}
    $\tilde{y}_x = \underset{1\leq i \leq k}{\mathrm{argmax}} (\zeta_i(x))$\;
    where, $\zeta_i(x) \coloneqq \sigma_i + D_i(x) $ 
   }
   \Comment{\teal{Pull loss: to reduce the distance from sample's own class and make compact clusters, }}
   $L_{\mathrm{P}} = \sum_{x \in B} D_{y_x}(x)$\; 
   \Comment{\teal{Score loss: to make sure score from sample's own class is non-negative and from rest classes it's negative, }}

 $L_{\mathrm{SL}} =
    \begin{cases}
    \sum_{x \in B} \sum_{i} \frac{\exp(\zeta_i(x))}{\#|B|} \hbox{ for } i \in \{1,2,\ldots, k\}\backslash \{y_x\}, \\
        \sum_{x \in B}\left( \mathrm{ReLU} (- \zeta_{y_x}(x)) + \mathrm{log}(1 +\zeta_{y_x}(x)^2 ) \right), \hbox{else}
    \end{cases}$\;
    
   \Comment{\teal{Distance and Score-based Effective Focal loss: }}
   $L_{\mathrm{EFL_1}} = \sum_{x \in B}\mathrm{EFloss}(y_x^o, 1/D(x))$\; \Comment{\teal{$y_x^o$ is one hot label vector for $x$}}
   $L_{\mathrm{EFL_2}} = \sum_{x \in B}\mathrm{EFloss}(y_x^o,\zeta_i(x))$\;
   \Comment{\teal{Net-loss}}
   $\mathcal{L}_{net}= L_{P} + L_{SL} + L_{EFL_1} + L_{EFL_2}$\;
   \Comment{\teal{Use Block Coordinate Descent (BCD) to update:}}
    (a) DNN's parameters viz $W$ and\;
   (b) Mean and standard deviation for k in-distribution classes viz $(\mu_i, \sigma_i)$ for $i \in \{1, 2, \ldots, k\}$\;
   }
   \Comment{\teal{Prediction for test data containing both in-distribution (ID) and OOD samples:}}
   \For{$x' \in \mathcal{D}_{test}$}
   {\begin{equation*} 
  \tilde{y}_x' =
    \begin{cases}
       OOD & \textbf{if } \zeta_i(x) < 0 \hbox{ for } i \in \{1,2, \ldots, k\}\\
      \underset{1\leq i \leq k}{\mathrm{argmax}} (\zeta_i(x)), & \textbf{otherwise}
    \end{cases}       
\end{equation*}}
\end{algorithm}
\label{our_algorithm}  
detection, will yield a performance boost when employing DNN-GDID. For effective focal loss we use \gamma value of 

\bibliographystyle{IEEEbib}
\bibliography{reference}